\documentclass[letterpaper]{article}
\usepackage{aaai}
\usepackage{times}
\usepackage{helvet}
\usepackage{courier}
\usepackage{graphicx}
\usepackage{booktabs}
\usepackage{multirow}
\usepackage{amsmath}
\usepackage{amssymb}
\usepackage{xcolor}
\usepackage{enumitem}
\usepackage{hyperref}
\newcommand{\openbiorq}{OpenBioRQ}

\title{OpenBioRQ: Unsolved Biomedical Research Questions for Agents}
\author{Minbyul Jeong \\
Upstage AI \\
minstar@upstage.ai \\
\href{https://huggingface.co/datasets/Minbyul/OpenBioRQ}{Dataset} \quad 
\href{https://github.com/minstar/healthcare-research}{Code}
}

\begin{document}
\maketitle

\begin{abstract}
A working citation looks like proof---but the fact that a link resolves does not mean the cited paper supports the claim.
I find that current agentic models rarely fabricate citations (over $99\%$ resolve), yet roughly $15.9\%$ link to the wrong paper.
Existing benchmarks miss this failure mode: when a question has a fixed answer key, a model can reproduce the expected source from that key rather than independently verifying that the source supports the claim.
I introduce \textbf{\openbiorq{}}, a retrieval-grounded agentic benchmark of $12{,}553$ unsolved biomedical research questions across $12$ domains that treats open questions as a faithfulness-and-abstention probe.
To my knowledge, this is the first biomedical benchmark to combine an agentic setting---where the model must issue multiple tool calls---with unsolved questions that have no answer key.
Openness is verified against real follow-up evidence rather than a model's parametric knowledge.
Difficulty is empirical: I anchor it on questions that three open-weight reference models fail to answer, rather than on subjective hardness labels.
On this hardest subset, held-out models from the same lineage as the difficulty anchors solve only $\sim$$17\%$, while three independent frontier agents (Gemini-3-Pro, Opus-4.7, GPT-5.5) span a wide $29$--$60\%$ range.
The benchmark is thus hard, non-saturating (the best agent still leaves $\sim$$33$--$40\%$ unsolved), and discriminating across capability tiers.
Beyond difficulty, I observe agentic collapse on the hardest questions, where agents stop using their tools.
For the most collapse-prone model, blocking tool access entirely barely changes its score---so tools stop paying off exactly where they are needed most.
A frozen per-question checklist raises inter-judge agreement from Spearman $0.35$ to $0.82$.
\openbiorq{} targets research assistance---evidence retrieval and faithful citation---not clinical decision support.

\end{abstract}

\section{Introduction}

When an agentic model answers a biomedical question and attaches a PMID, the resolvable identifier
reads as evidence: a reader follows the link, the record exists, and the claim inherits the authority
of the literature. For example, asked about the long-term safety of mRNA vaccines, a model reports
that ``vaccine efficacy remained $\sim$$91\%$ against symptomatic COVID-19'' and cites PMID~\href{https://pubmed.ncbi.nlm.nih.gov/34407296/}{34407296}.
The identifier resolves to a real, indexed article---but an ophthalmology study that never mentions a
vaccine. The citation exists, yet it does not support the claim. The failure is purely one of linking
the claim to its source: the reported efficacy is roughly correct and the model even names the right
trial in prose---only the emitted identifier points elsewhere. The example is a verbatim trace from
my audit, not a hypothetical.

\begin{figure}[t]
\centering
\includegraphics[width=\columnwidth]{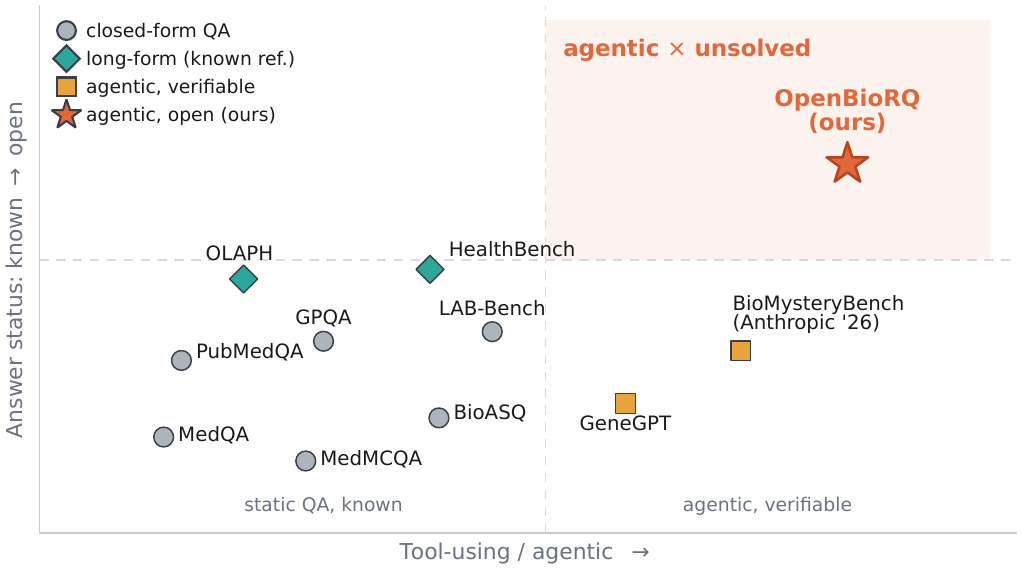}
\caption{\textbf{Where \openbiorq{} sits among biomedical benchmarks}: answer status
(known $\rightarrow$ genuinely open) against tool use (static $\rightarrow$ agentic). Prior
benchmarks---closed-form, long-form, and agentic-with-answers (e.g., BioMysteryBench)---all assume a
known answer. \openbiorq{} alone targets unsolved questions, in the shaded
\emph{agentic $\times$ unsolved} quadrant, where wrong-paper citation, agentic collapse, and
abstain-vs-confabulate become measurable.}
\label{fig:motivation}
\vspace{-0.3cm}
\end{figure}

Closed-form question answering~\cite{tsatsaronis2015bioasq,jin2019pubmedqa,jin2021medqa,pal2022medmcqa}
requires models to answer solvable questions, where the correct answer---and often its supporting
source---is fixed in advance. Long-form question
answering~\cite{abacha2017overview,abacha2019bridging,singhal2023medpalm,manes2024k,jeong2024olaph}
relaxes the answer format to free text, but still scores each response against a fixed gold answer and
its curated reference, so faithfulness is judged by overlap with a pre-specified source rather than by
whether the model's own citations actually support its claims. Recently, agentic evaluations such as
GeneGPT~\cite{jin2024genegpt} and BioMysteryBench~\cite{biomysterybench2026} let models call external
tools and gather evidence on their own, yet they still target questions with a verifiable ground-truth
answer, leaving unchecked whether each retrieved paper actually supports the claim it is attached to.

These three families differ in format but share a single assumption, a known answer, and so occupy
three corners of an answer-status $\times$ agentic plane (Figure~\ref{fig:motivation}). The fourth
corner is underexplored, and the reason to target it is structural, not aesthetic. The failures I
studied---wrong-paper citation, agentic collapse, and the choice between abstaining and
confabulating---surface only when the answer key is removed. When the answer is fixed in advance, a
model can pass by echoing the source it was handed, and these failures have no room to appear. Thus,
\textbf{\openbiorq{}} targets: $12{,}553$ \emph{unsolved} biomedical research questions across $12$
domains. I pose them not to be answered against a key, but to be attempted---and I score how an agent
behaves in the attempt. To my knowledge, it is the first benchmark in the agentic $\times$ unsolved
regime. Reframing the task this way turns evaluation into a faithfulness-and-abstention probe.

I audit every citation an agent emits at two levels: (1)~existence, whether the identifier resolves to
a real paper; (2)~content support, whether that paper actually supports the claim. One might expect
these to rise and fall together. In biomedical research, they come apart and are opposite to what
other fields report. ResearchMath~\cite{researchmath14k} finds almost $54\%$ of references in
mathematical writing to be outright fabricated; biomedical agents rarely fabricate, with only $0.7\%$
of $4{,}863$ emitted citations failing to resolve. Of the citations that do resolve, $15.9\%$ are
wrong-paper that does not support the claim it is attached to (an LLM-judge estimate; $10.6\%$ under a
different-family judge, \S\ref{sec:experiments}). Because a wrong-paper citation resolves, a reader is
more inclined to trust it; it may therefore be more hazardous than an obvious fabrication---a severity
gap I pose as a hypothesis.

Scoring unsolved questions invites an immediate objection: with no key, how do we know a question is
genuinely open, and that it is genuinely hard? Three design choices answer this rather than assume it.
(1)~Framing a question as open is not enough---doing so alone collapses into confirmation bias.
Instead, a retrieval-grounded verifier re-judges each question against real follow-up evidence, and a
re-audit of the $657$-question core finds none of them resolved; (2)~the core retains only questions
the open-weight reference roster jointly fails, and on this subset held-out models of the same lineage
solve just $17\%$ while three independent-lineage frontier agents span $29$--$60\%$---hard at the
open-weight tier, yet non-saturating and discriminating, since even the strongest agent leaves over a
third unsolved; (3)~answers are produced through multi-round tool use over ten biomedical REST APIs and
graded against a frozen per-question checklist, which raises inter-judge agreement from Spearman $0.35$
to $0.82$ (\S\ref{sec:discussion}).

Beyond citations, \openbiorq{} surfaces an agentic collapse: on the hardest questions, agents abandon
their tools, and for the most collapse-prone model, blocking tool use altogether barely changes
performance---tool access stops paying off exactly where it should matter most~(\S\ref{sec:experiments}).
\openbiorq{} is a research-assistance benchmark for evaluating literature synthesis and grounding, not a
clinical decision-support tool, and must not be used to guide patient care.

\section{Related Work}
\label{sec:related}

\paragraph{Biomedical question answering benchmarks.}
The closed-form medical QA that trained a generation of models is now largely exhausted as a
discriminator. On MedQA~\cite{jin2021medqa} and its
kin~\cite{tsatsaronis2015bioasq,jin2019pubmedqa,vilares2019headqa,hendrycks2021mmlu,pal2022medmcqa},
frontier systems---from the Med-PaLM line~\cite{singhal2023medpalm,singhal2023medpalm2}
onward---cluster so tightly that a recent evaluation finds specialized clinical tools and general
LLMs statistically indistinguishable, clustered in an $\sim$$8$-point band at
$88$--$96\%$~\cite{vishwanath2026generalist}. Harder, expert-written sets
(GPQA~\cite{rein2023gpqa}, LAB-Bench~\cite{laurent2024labbench}) and even agentic biomedical
evaluation (BioMysteryBench~\cite{biomysterybench2026}) raise the ceiling, but they raise it within
the same regime: every question still has a known, verifiable answer. BioMysteryBench makes the
assumption explicit, choosing problems whose answers are verifiable by design.

What \openbiorq{} inherits from this lineage is its scoring machinery, not its answer model.
The closest precedent decomposes a free-text answer into reusable rubric statements: long-form
MedLFQA/OLAPH grades each response against reference must-have/nice-to-have
sentences~\cite{jeong2024olaph}, and HealthBench against physician-written per-conversation
criteria~\cite{arora2025healthbench}. Both still presuppose an answer---the rubric measures how
closely a response matches a curated reference. \openbiorq{}'s frozen per-question checklist is a
direct descendant of that decomposition, carried to unsolved questions (JLA priority
partnerships, NICE guidance, the open-question literature) that have no consensus answer at
construction time; I keep the decomposed-rubric machinery but change what it scores---from
\emph{correctness against a reference} to \emph{grounding and acknowledged uncertainty}
(\S\ref{sec:evaluation}).

\paragraph{Research-level and open-problem datasets.}
My construction methodology most directly extends ResearchMath-14k~\cite{researchmath14k}: I adopt
its agentic extract-then-refine pipeline, its LLM-judge self-containment audit (\openbiorq{} attains
$85.4\%$ self-contained---$95\%$ on the retrieval-verified track, $75\%$ on expert-consensus;
Appendix~\ref{app:construction}), its MiniLM-L6 near-duplicate removal at cosine $0.90$, and its
separation of behavioral from factuality characterization. I then extend it in two
ways the open biomedical setting forces. First, openness cannot be taken on faith, so \openbiorq{} is
retrieval-grounded: source-framing-only judgment collapsed to zero
\texttt{answered}/\texttt{unknown} labels (a confirmation bias), and a retrieval-grounded status
verifier instead re-confirms each core question against real follow-up evidence (none of the $657$ met
my resolution criterion). Second, where ResearchMath finds a ``generation paradox'' of
fabricated references, the biomedical agents I evaluate barely fabricate at all, so I replace
its single existence check with a two-level audit that continues to content support
(\S\ref{sec:experiments}) and exposes real papers that do not support the claim.

\paragraph{Tool-using and agentic LLM evaluation.}
The multi-round paradigm I adopt was established by ReAct-style reasoning--action
interleaving~\cite{yao2023react} and learned tool invocation~\cite{schick2023toolformer}, and a wave
of general agentic benchmarks now scores whether a model can select, parameterize, and sequence tools
to finish a
task~\cite{patil2023gorilla,mialon2023gaia,qin2024toolllm,yao2024taubench,trivedi2024appworld,wei2025browsecomp,jeong2026healthcare}.
What unites them is a verifiable target: a run is correct only if it reaches the known answer or
end-state. The same holds in medicine, where AgentClinic~\cite{schmidgall2024agentclinic} casts
diagnosis as sequential agentic dialogue and MedAgentBench~\cite{jiang2025medagentbench} scores tool
use against a FHIR-compliant EHR---each grading against a correct diagnosis or a completed task.
\openbiorq{} removes that target: with no verifiable answer, what I measure becomes
grounding and abstention---the behavior actually correct on an unsettled
question---which ties my setting to selective prediction and calibrated
abstention~\cite{kamath2020selective,kadavath2022know}. I use tools differently from
GeneGPT~\cite{jin2024genegpt}, which couples an LLM to NCBI endpoints to improve access: the
same ten-tool REST stack becomes for me an evaluation instrument, which is what lets me observe
agentic collapse (\S\ref{sec:experiments})---a failure mode answer-only benchmarks cannot see.

\paragraph{Citation faithfulness, attribution, and grounding.}
A line of work has measured how, and how often, model citations go wrong: that LLMs fabricate
plausible-looking references~\cite{walters2023fabricated}, whether a generated statement is
attributable to its cited source~\cite{bohnet2022attributed,rashkin2023attribution}, and how to make
systems cite retrieved evidence inline~\cite{gao2023ragfaithfulness}. Closest to my L2 question,
prior work audits whether the inline citations of generative search engines actually support
their statements---citation precision and recall, by human raters on general-domain
queries~\cite{liu2023verifiability}. I operationalize that question over the citations agents
themselves emit during multi-round tool use, rather than a curated statement--source set, and
at scale in biomedicine---where the picture inverts the math and general-domain findings above:
fabrication is near-zero ($\approx 0.7\%$), while real-but-unsupporting citation is the dominant
failure. I propose no new attribution metric; the contribution is this empirical inversion
(\S\ref{sec:experiments}), with the severity gap posed as a hypothesis (Appendix~\ref{app:discussion}).

\section{The \openbiorq{} Benchmark}
\label{sec:benchmark}

One extracted fragment reads only ``the pathophysiology of PSVD remains unclear'' (PMID~40578802);
\openbiorq{}'s refine pass turns it into a stand-alone question---``What is the pathophysiologic
mechanism of portosinusoidal vascular disorder (PSVD)?''---then attaches the three layers that make an
unsolved question gradable. \openbiorq{} is a corpus of such questions, each carrying a
retrieval-grounded openness label (is it actually still open?), an empirical difficulty
signal (do current models fail it?), and a frozen per-question checklist rubric (how is an
answer scored?). The
protocol that consumes those layers is deferred to \S\ref{sec:evaluation}; here I describe how
the corpus is built and audited. Construction runs a
\textsf{crawl}\,$\rightarrow$\,\textsf{extract}\,$\rightarrow$\,\textsf{refine}\,$\rightarrow$\,\textsf{dedup}\,$\rightarrow$\,\textsf{export}
pipeline, and the two label layers are then added on top of the exported text rather than baked into
it, so each can be re-audited independently. All statistics below are for the text-deduplicated
corpus; the pipeline mechanics, the rejected preprint source, worked example items with their
checklists, and per-category counts are in Appendix~\ref{app:construction}.

\subsection{Tracks}
\label{sec:tracks}
Every question records the \texttt{corpus\_track} it came from, because the track is why I
call it open (Table~\ref{tab:tracks}): \texttt{retrieval\allowbreak\_verified} questions
(from PubMed, trial registries, and arXiv) are open because real follow-up evidence says so;
\texttt{expert\_consensus} questions are open because an authoritative body has declared them
so---JLA Priority Setting Partnerships~\cite{jla} and NICE research recommendations~\cite{nice};
\texttt{priority\_setting} draws on society agendas and WHO/CHNRI/NASEM/PCORI and Delphi documents via
Europe PMC~\cite{europepmc}; and \texttt{expand} draws on Cochrane~\cite{cochrane} research-gap
literature.

\begin{table}[t]
\centering
\small
\setlength{\tabcolsep}{3pt}
\begin{tabular}{@{}llr@{}}
\toprule
Track & Source & Count \\
\midrule
\texttt{retrieval\_verif.} & PubMed / trials / arXiv & $6{,}648$ \\
\texttt{expert\_consensus}   & JLA PSPs, NICE recs & $5{,}905$ \\
\texttt{priority\_setting}   & WHO/CHNRI/NASEM, Delphi & $525$ \\
\texttt{expand}              & Cochrane research gaps & $483$ \\
\bottomrule
\end{tabular}
\caption{The four \openbiorq{} tracks and their counts. The
\texttt{retrieval\allowbreak\_verified} and \texttt{expert\_consensus} tracks form the
$12{,}553$-question base corpus; the \texttt{priority\_setting} and
\texttt{expand} tracks are additional and deduplicated against it.}
\label{tab:tracks}
\end{table}

\subsection{Construction Quality and Grounded Openness}
\label{sec:construction-quality}\label{sec:grounded}

\paragraph{Self-containment.} A question is only useful in isolation if it reads without its source
document---the PSVD rewrite is exactly that move---so I audit it directly: an LLM-judge check
($n{=}500$) finds $85.4\%$ of refined questions self-contained (\texttt{retrieval\allowbreak\_verified} $95\%$,
\texttt{expert\_consensus} $75\%$). The gain comes from the refine pass, as a before/after ablation
shows---an extractor-only baseline is only $51.6\%$ self-contained, so refinement adds $+33.8$
points. Near-duplicates are then collapsed by a MiniLM-L6~\cite{minilm} cosine~$\geq 0.90$ screen over
question text, which keeps distinct questions extracted from the same source paper (audit
detail in Appendix~\ref{app:construction}).

\begin{figure*}[t]
\centering
\includegraphics[width=0.92\linewidth]{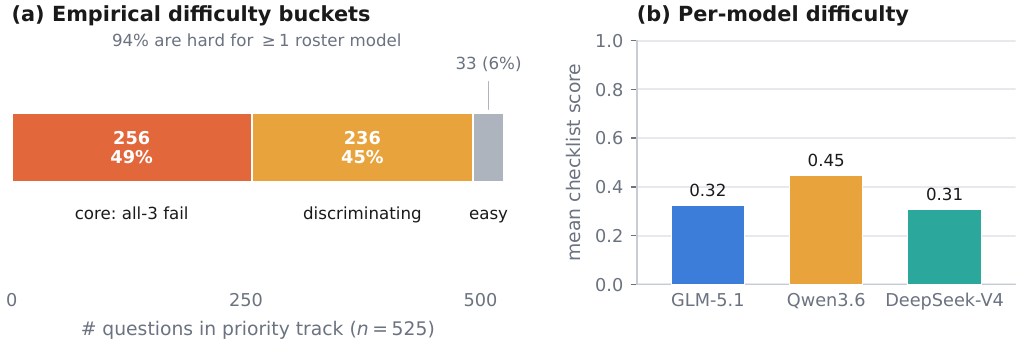}
\caption{Empirical difficulty on the question-granular \texttt{priority\_setting} track
($n{=}525$). \textbf{(a)} bucket composition ($\sim$49\% core / 45\% discriminating / 6\% easy);
\textbf{(b)} mean checklist score per roster model---all three fall below the $0.5$ pass threshold.}
\label{fig:difficulty}
\vspace{-0.2cm}
\end{figure*}

\paragraph{Retrieval-grounded openness.} Openness is judged from real follow-up evidence, not from a
model's memory of how its source framed the problem---and the distinction is not cosmetic:
source-framing-only refinement produced complete confirmation bias, labeling zero questions
\texttt{answered} or \texttt{unknown}. The fix is to re-decide status only from evidence the model
actually gathered. A retrieval-grounded Stage-2 judge does exactly this---citing evidence IDs, or else
falling back to \texttt{unknown}---which makes resolution detectable at last: it changes the status of
$56.5\%$ of a $200$-question sample and reaches \texttt{unknown} for $14\%$. Run over the hardest
subset, a still-open re-audit of the $657$ core questions finds \textbf{$0/657$} with
detectable resolution (narrative or guideline resolution is undetectable by this audit, and its
recall is unquantified). That the detector is not simply blind is confirmed by a contamination
positive control---$34$ injected known-contaminated items---which it catches at $100\%$ recall and
$0\%$ false-positive. Verbatim- and perplexity-based membership checks add a further negative: no
memorization signature (all four audits in Appendix~\ref{app:construction}).

\subsection{Empirical Difficulty}
\label{sec:difficulty}
\begin{figure}[t]
\centering
\includegraphics[width=0.92\linewidth]{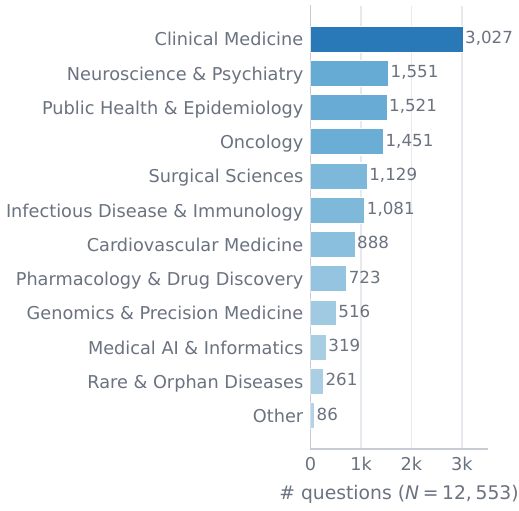}
\caption{Domain coverage: the $12$ top-level taxonomy categories over the \openbiorq{} corpus
($12{,}553$ questions); per-category counts are labelled directly on the bars.}
\label{fig:taxonomy}
\end{figure}

Difficulty is assigned empirically rather than by hand: every question is answered with full
tool access by three roster models---GLM-5.1~\cite{glm}, Qwen3.6~\cite{qwen}, and
DeepSeek-V4~\cite{deepseek}---and graded against its frozen checklist. A question is
core exactly when all three fail it (checklist score $<0.5$). On the cleanest track
(\texttt{priority\_setting}, labeled at question granularity end-to-end), this yields $49\%$ core,
$45\%$ discriminating, and only $6\%$ easy (Figure~\ref{fig:difficulty}a); the three roster models
all sit below the $0.5$ pass threshold in mean checklist score (Figure~\ref{fig:difficulty}b), so the
all-fail bucket is a property of the open-weight tier rather than of any single model. Across the full
corpus the $12{,}553$ questions spread over $12$ level-one taxonomy domains with no domain dominating
(Figure~\ref{fig:taxonomy}); per-domain difficulty and citation breakdowns are deferred to
Appendix~\ref{app:extended}.

\section{Evaluation Protocol}
\label{sec:evaluation}

Consider an open question \openbiorq{} must grade---``Can therapies targeting the glymphatic system
prevent or slow Alzheimer's disease?'' Its frozen checklist requires an answer to
\texttt{must\_mention} AQP4 polarization as a mechanistic target, \texttt{must\_acknowledge} that no
glymphatic-targeting therapy has entered an AD trial, and \texttt{must\_avoid} claiming any such
therapy is proven---so grading reduces to a handful of concrete, checkable verdicts. But evaluating
genuinely unsolved questions is itself an open problem, for two reasons that compound. First, there is
no single correct answer to grade against. Second, the qualities that matter on an open
question---evidentiary grounding, calibrated uncertainty, and the avoidance of confident
fabrication---are precisely the ones brittle string- or embedding-overlap metrics cannot see. \openbiorq{} therefore couples two pieces: an agentic completion harness,
in which a model actively gathers evidence through biomedical tools, and the per-question frozen
checklist just illustrated, which converts open-ended grading into a set of concrete, reproducible
verdicts (the end-to-end flow is in Appendix~\ref{app:formalism}, Figure~\ref{fig:evalflow}).

\subsection{Checklist Scoring}
\label{subsec:notation}\label{subsec:checklist}
The natural first choice---asking a strong LLM to score each answer along a few abstract dimensions on
a continuous scale~\cite{zheng2023judging,liu2023geval}---is unreliable here, and for a specific
reason: with no fixed referent, different judges (and the same judge across runs) anchor those
dimensions differently, a known hazard of LLM-as-judge protocols~\cite{zheng2023judging} that
motivates structured, rubric-based grading~\cite{liu2023geval}. The effect is measurable---replacing
free-form dimension scoring with the frozen checklist below raised inter-judge (LLM-vs-LLM)
agreement from a Spearman correlation of $0.35$ to $0.82$ (human-expert agreement deferred; see
Limitations).

The core move is to lift grading from vague dimensions to a per-question checklist generated
once and then frozen. For each question $q$, a strong model drafts five to eight concrete, checkable,
question-specific criteria $C_q=\{(t_i,w_i)\}_{i\in I_q}$, each with a type $t_i$ and an importance
weight $w_i\in\{1,2,3\}$:
\begin{itemize}[leftmargin=1.4em,itemsep=1pt,topsep=2pt]
  \item \textbf{must\_mention} --- a key fact, mechanism, or method the answer should include;
  \item \textbf{must\_acknowledge} --- an uncertainty, gap, or open aspect it must flag (critical for
    unsolved questions);
  \item \textbf{must\_ground} --- a claim it must back with real cited evidence (PMID, trial, or tool
    result);
  \item \textbf{must\_avoid} --- a behavior that should not occur (asserting a definitive
    answer to an open question, fabricating citations, or falsely claiming the tools returned
    nothing).
\end{itemize}
A judge reads only the answer text and its tool calls and assigns each criterion a verdict
$v_i\in\{1,0.5,0\}$ (met/partial/not\_met; the partial value is the scale midpoint, not a tuned
parameter). For \texttt{must\_avoid} the polarity is inverted---$v_i{=}1$ means the prohibited
behavior was correctly avoided---so a higher score can never reward confabulation. The question score
is
\begin{equation}
  \mathrm{score}(q)=\frac{\sum_{i\in I_q} w_i v_i}{\sum_{i\in I_q} w_i}\in[0,1],
  \label{eq:checklist}
\end{equation}
bounded and individually auditable.

A model $m$ \emph{solves} $q$ iff $\mathrm{score}^T_m(q)\ge0.5$ under a single deterministic
($T{=}0$, one attempt) completion---my headline \textbf{solve rate}. I state the decoding
temperature wherever it is used, because difficulty turns out to be decoding-sensitive
(\S\ref{sec:exp:core}). The $0.5$ cutoff is only a reporting convention, not a load-bearing
threshold: my claims rest on the capability ordering and the cross-tier band, both
fixed by the underlying mean scores ($0.39/0.44/0.53$ for the frontier agents) and therefore robust
to where the cutoff is drawn.

Freezing the checklist before grading is what drives the agreement gain, because every judge then
evaluates the same criteria; it also makes the metric robust to agentic collapse, since
a collapsed or hedging answer fails its \texttt{must\_mention}/\texttt{must\_ground} criteria rather
than earning credit for fluent prose. The full notation---the temperature-parametric failure count $F_T$, the
core and frozen-core set-builders, the citation fetch-outcome algebra, and the
openness-status constraint---is in Appendix~\ref{app:formalism}.

\subsection{Agentic Harness and Gold Answers}
\label{subsec:harness}\label{subsec:goldnotgt}
Each question is answered through multi-round tool use rather than a single forward pass: a model
receives the question and access to ten biomedical retrieval tools wrapping public REST
APIs---\texttt{pubmed}, \texttt{clinicaltrialsgov}, \texttt{openfda}, \texttt{opentargets},
\texttt{chembl}, \texttt{uniprot}, \texttt{pubchem}, \texttt{kegg}, \texttt{ncbi\_datasets},
\texttt{biomcp}---and may interleave reasoning and tool calls for up to ten rounds before committing
to a final answer. Each trace records the final answer, the ordered tool calls, token usage, and
wall-clock time, and every question carries a unique \texttt{task\_id} of the form
\texttt{source\_id\#k}, so multiple questions from one source paper are scored independently. The
wrong-paper verdict the central result rests on is an
LLM-cross-family proxy; agreement with human medical experts is deferred to future work, and I
release a human-annotation set to enable it.

Each question also comes with an LLM-synthesized ``gold'' answer, but I use it only as
reference context when drafting rubrics---never as a target to match or a source of ground-truth
citations. The reason is empirical, and it is the very hazard the paper studies: $\approx 100\%$ of
the gold answers' cited PMIDs resolve, yet $\approx 74\%$ are gold-misattributions (the cited
work does not support the claim; LLM-judge, $n\approx360$, two-judge robust). Treating those citations
as ground truth would propagate plausible-looking but unsupported attributions straight into the
metric, so I instead assess grounding through the \texttt{must\_ground} criteria, which require the
model's own answer to cite genuinely supporting evidence (full sub-finding in
Appendix~\ref{app:discussion}).

\section{Experiments and Analysis}
\label{sec:experiments}

I organize the study around three questions. First---the heart of this paper---when these agents
cite the literature, are the citations real, and do the real ones support the claims they are
attached to (\S\ref{sec:exp:citation})? Second, how do the agents behave on open biomedical
questions, and how much does that diverge across models (\S\ref{sec:exp:behavior})? Third, how hard
is the empirically-derived core set---for the roster that defined it and for held-out frontier
agents---and how sensitive is that hardness to decoding (\S\ref{sec:exp:core})? Unless noted,
behavioral and citation analyses run over the $1{,}969$-question gold-answer slice (the L1/L2 audit
covers the $4{,}863$ citations the three roster models emit across their trajectories---the gold
slice plus the priority and expand tracks), and core-set results over the $657$-question core (with a
$423$-question frozen-core subset defined at $T{=}0$). All agentic runs share one harness (ten-tool,
frozen-checklist; \S\ref{sec:evaluation}); Figure~\ref{fig:overview} previews the two headline
results. Construction and contamination audits are in Appendix~\ref{app:construction}; extended
breakdowns, robustness checks, and the per-domain table are in Appendix~\ref{app:extended}.

\begin{figure*}[t]
\centering
\includegraphics[width=\linewidth]{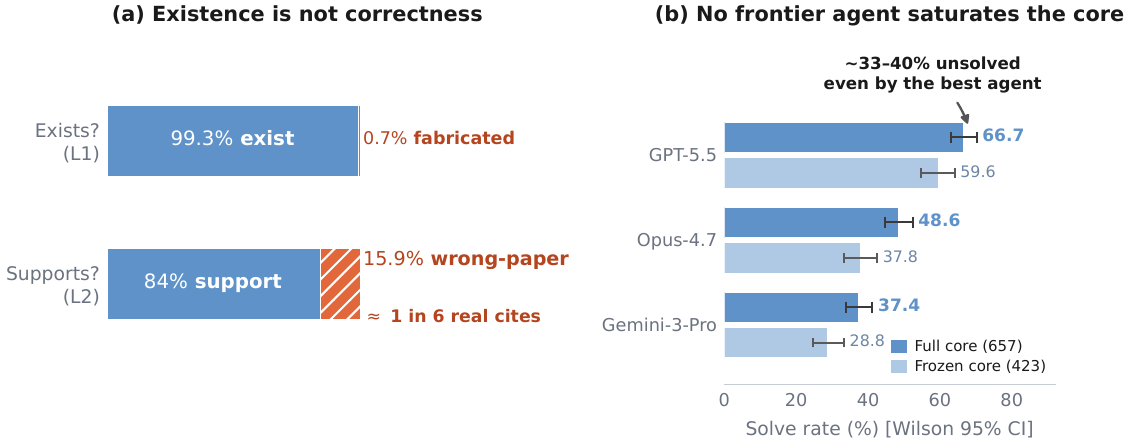}
\caption{The two headline results. \textbf{(a)} Existence is not correctness: almost every citation
exists (L1), yet $15.9\%$ of the real ones are wrong-paper citations that do not
support their claims (L2)---an existence-only check is false comfort (per-model breakdown in
Figure~\ref{fig:citation}). \textbf{(b)} No frontier agent saturates the core: independent-lineage
agents span a wide solve-rate range on the frozen ($423$) and full ($657$) core (Wilson $95\%$
CIs), and even the best leaves a third or more unsolved.}
\label{fig:overview}
\vspace{-0.2cm}
\end{figure*}

\subsection{Two-Level Citation Factuality}
\label{sec:exp:citation}

\begin{figure*}[t]
\centering
\includegraphics[width=\linewidth]{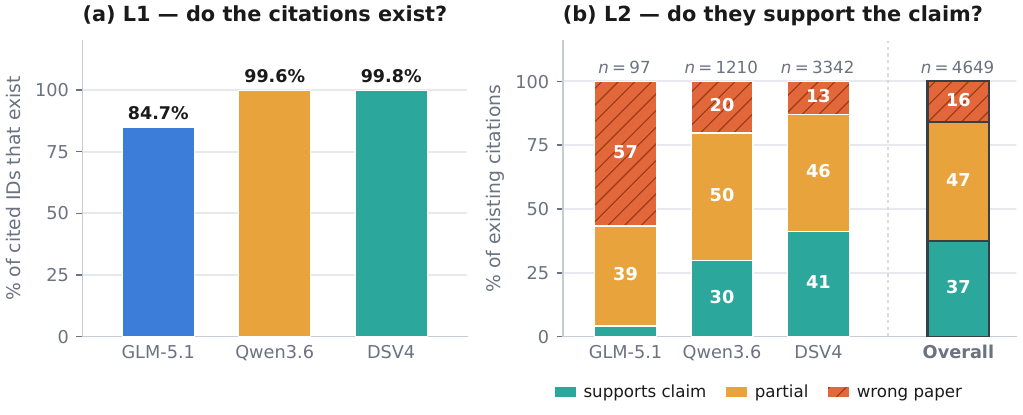}
\caption{The two-level citation audit at full per-model resolution (Figure~\ref{fig:overview}a pools
these). \textbf{L1 (left):} emitted citations almost all exist, with GLM-5.1 the exception
(fabricating mostly trial NCT identifiers). \textbf{L2 (right):} of the real citations, the
fraction whose cited paper supports the local claim (green=yes, orange=partial,
red\,(hatched)=wrong-paper). Existence $\neq$ correctness.}
\label{fig:citation}
\end{figure*}

Prior work asks the citation question too narrowly.
Work on fabricated references in mathematical and general scientific
writing~\cite{walters2023fabricated,researchmath14k} treats a citation as a binary existence
problem---does the identifier resolve? That test is necessary, but in biomedicine it is badly
insufficient: a real PMID attached to a wrong clinical claim is a more consequential
misattribution than an obviously fake one, precisely because the identifier resolves and a downstream
reader is more likely to trust it (a sourcing argument; I make no patient-impact or claim-veracity
claim). I therefore audit at two levels (Figure~\ref{fig:overview}a; full per-model breakdown in
Figure~\ref{fig:citation}): \textbf{L1, existence} (does the identifier resolve?) and
\textbf{L2, content support} (does the cited record support the claim?).

\paragraph{L1 --- existence.}
A naive existence check badly overcounts fabrication by more than an order of magnitude---my first
pass mis-scored transient rate-limit failures as missing records---so I use a \textbf{tristate fetch}
that counts only genuinely empty records as fabricated and re-checks transient failures. On the corrected audit over $4{,}863$ citations,
biomedical agents almost never invent identifiers: Qwen3.6 reaches $99.6\%$ existence and DeepSeek-V4
$99.8\%$; GLM-5.1 is the exception at $84.7\%$ (Wilson $95\%$ CI $[78.2,89.5]$), its fabrications
concentrated in trial NCT identifiers emitted ``based on training knowledge.'' Aggregated, only
$\approx 35$ of $4{,}863$ identifiers are fabricated---$\mathbf{0.7\%}$, the opposite of the
$\sim$54\% fake-reference rate reported for ResearchMath~\cite{researchmath14k} (per-model L1 in
Figure~\ref{fig:citation}). An L1-only audit would conclude---incorrectly---that citation
quality is excellent. \textbf{Existence is not correctness.}

\paragraph{L2 --- content support and the wrong-paper failure.}
The real story lives at L2. For every resolving citation, my primary GLM-5.1 judge reads the cited record's
title and abstract and decides whether it \texttt{supports}, \texttt{partially supports}, or does
\texttt{not} support its claim; a real identifier that does not support is a \textbf{wrong-paper}
citation---PMID~34407296 is one such. After correcting a claim-extraction artifact (below), over $4{,}649$ real citations
(Table~\ref{tab:l2}) \textbf{$15.9\%$ are wrong-paper} under this primary judge. The pattern
inverts L1:
DeepSeek-V4 and Qwen3.6, which almost never fabricate, are nonetheless wrong on $13.1\%$ and $20.2\%$
of their real citations---so the wrong-paper finding survives dropping the small-sample GLM-5.1
($n{=}97$; Table~\ref{tab:l2}) entirely. The $15.9\%$ is micro-averaged (DeepSeek-V4
dominates the pool; the macro-average is a higher $\approx 30.0\%$, inflated by GLM-5.1's rate).

\paragraph{The wrong-paper rate is robust to the judge.}
I re-judged the entire L2 population with an independent different-family judge (Opus-4.7).
Both rank GLM-5.1 worst and DeepSeek-V4 best, and agree on the binary verdict at Cohen's
$\kappa=0.755$ ($n{=}118$). One artifact had to be corrected first: a claim-extraction bug left $\sim$11\% of snippets
as bare reference fragments, which a judge with no claim to assess defaulted to non-support; after
re-extracting and re-judging those under both judges (themselves wrong-paper $20.4\%$ either way),
the definitive rates are $15.9\%$\,[14.9,\,17.0] (primary GLM, $n{=}4{,}649$) and
$10.6\%$\,[9.7,\,11.5] (Opus, $n{=}4{,}745$)---two judge families, not a single-judge artifact. The
estimate is bracketed on both sides: a full-text re-judge ($16.1\%$, $n{=}149$) runs higher
than abstract-only, and counting the large ``partial'' bin ($28.1\%$) as non-support would push it
higher still. Conversely, a preliminary non-expert human spot-check flags fewer, agreeing on
the clear cases but only at binary $\kappa{=}0.29$/$0.51$ (vs the LLM-vs-LLM $0.755$). I therefore
read $15.9\%$ as a judge-relative estimate pending expert adjudication
(Appendices~\ref{app:extended},~\ref{app:discussion}).

\begin{table}[t]
\centering
\small
\begin{tabular}{@{}lrrrcc@{}}
\toprule
\textbf{Model} & \textbf{Real} & \textbf{Supp.} & \textbf{Part.} & \multicolumn{2}{c}{\textbf{Wrong-paper \%}} \\
\cmidrule(lr){5-6}
               & \textbf{cites} & \textbf{\%} & \textbf{\%} & \textbf{GLM} & \textbf{Opus} \\
\midrule
GLM-5.1      & $97$      & $4.1$  & $39.2$ & $56.7$\textsuperscript{\dag} & $64.9$ \\
Qwen3.6      & $1{,}210$ & $29.8$ & $49.9$ & $20.2$ & $13.0$ \\
DeepSeek-V4  & $3{,}342$ & $41.2$ & $45.8$ & $13.1$ & $7.6$ \\
\midrule
Overall      & $4{,}649$ & $37.4$ & $46.7$ & $\mathbf{15.9}$ & $\mathbf{10.6}$ \\
\bottomrule
\end{tabular}
\caption{\textbf{L2 content-support audit on resolving identifiers.}
\emph{Wrong-paper} $=$ the identifier exists but does not support the claim.
Two independent judges (GLM-5.1, Opus-4.7) read title$+$abstract and rank GLM-5.1 worst \emph{despite} its near-perfect existence; rates are LLM-judge estimates ($15.9\%$/$10.6\%$ overall).
\textsuperscript{\dag}$n{=}97$, small-sample-fragile.}
\label{tab:l2}
\vspace{-0.2cm}
\end{table}

\paragraph{Why L2 is the consequential level.}
A wrong-paper citation is a misattribution that surface-citation~\cite{gao2023ragfaithfulness} and
fabrication~\cite{walters2023fabricated} audits cannot catch. The defect is one of linking a claim to its source,
not reasoning: a wrong-paper citation shows no detectable association with whether the answer is
otherwise grounded (\texttt{must\_ground}; independence not refuted rather than demonstrated;
Appendix~\ref{app:extended}). The citation is the audit trail a
reader follows, and a resolvable-but-unsupporting PMID silently breaks it even where the answer looks
well-grounded---so \openbiorq{} treats L2 content support as the citation metric of record.

\subsection{Behavioral Analysis of Tool-Using Agents}
\label{sec:exp:behavior}
\begin{figure}[t]
\centering
\includegraphics[width=\linewidth]{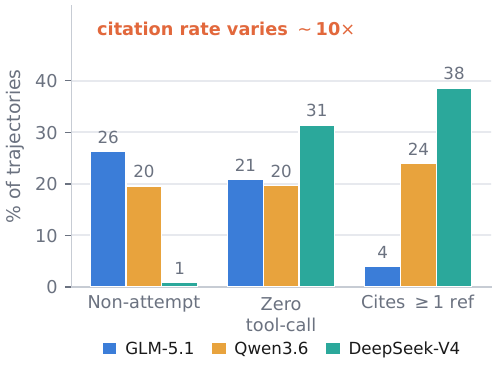}
\caption{Behavioral divergence across the three roster models on the $1{,}969$ set: non-attempt rate, agentic collapse rate, and citation rate. Behaviors differ by up to $\sim$10$\times$; on the harder priority track both GLM-5.1 and DeepSeek-V4 collapse to near-zero tool use while Qwen3.6 stays engaged.}
\label{fig:behavioral}
\end{figure}

The agents diverge as sharply in how they engage as in what they cite. Over the
$1{,}969$-question set (Figure~\ref{fig:behavioral}; full per-model table in
Table~\ref{tab:behavior}, Appendix~\ref{app:extended}) I measure the non-attempt rate, the
zero-tool collapse rate (an answer with no tool call---parametric recitation), average tool calls,
and the cite-rate. Even here the models split: DeepSeek-V4 almost always attempts ($0.8\%$
non-attempt) but collapses to zero tools most often ($31.3\%$); GLM-5.1 issues the most tool calls
($12.6$) yet has a far higher non-attempt rate ($26.2\%$); and the cite-rate spans $\sim$10$\times$ (GLM $3.9\%$ vs
DeepSeek-V4 $38.5\%$).

The split widens on the harder, question-granular priority track, where agentic
collapse appears in \textbf{two of the three roster models---two independent lineages}. GLM-5.1's
non-attempt rate jumps to $\mathbf{69\%}$ (zero-tool $65\%$) and DeepSeek-V4's to $\mathbf{62\%}$
(zero-tool $62\%$), each abandoning its tools to decline or recite from memory, while Qwen3.6 stays
engaged (zero-tool $22\%$, flat). That split makes the collapse neither a single-model
pathology nor an artifact of the questions alone---a model-divergent failure that final-answer
benchmarks~\cite{researchmath14k} cannot see (per-model breakdown in Appendix~\ref{app:extended}).

\subsection{Agentic Baseline on the Core Set}
\label{sec:exp:core}
How hard is the core for frontier tool-using agents---the roster that defined it and held-outs that
did not? The buckets were defined at $T{=}0.3$, where all three roster models scored near $0.3$ and
solve rate was effectively zero; re-measuring the same core at $T{=}0$ flips it: the full
$657$-question core becomes much more solvable (GLM-5.1 $26.6\%$). \textbf{Empirical
difficulty is decoding-sensitive}; the ``$\approx0\%$'' was a $T{=}0.3$ artifact. I therefore define
the \textbf{frozen core}: the $423$ of $657$ questions all three roster models still fail at $T{=}0$
(Table~\ref{tab:core}).

\paragraph{Hard and discriminating, not an absolute ceiling.}
Same-lineage open-weight held-outs---GLM-5~\cite{glm5} and Qwen3.5-397B~\cite{qwen35_397b}---solve only $\sim$17\% ($16.6\%$, $16.8\%$; overlapping CIs), and an
older Qwen3-235B~\cite{qwen3_235b} just $2.1\%$, so the hardness is not a single-roster artifact.
The stronger test is independent-lineage frontier agents (Gemini-3-Pro, Opus-4.7, GPT-5.5),
which reveal a capability range---$28.8\%$, $37.8\%$, $59.6\%$ (ordering and $\sim$31-point spread
preserved on the churn-stable subset; \S\ref{sec:exp:repro}). The picture is sharper than
``$\sim$17\%'': the frozen core is hard at the roster's open-weight tier, yet non-saturating
and discriminating---a strong frontier agent solves a majority while even the best leaves $\sim$40\%
unsolved. Because all three independent lineages find it non-trivial, the circularity concern is
substantially addressed, though the set is by construction selected on roster failures.

\paragraph{Tools confer no measurable advantage.}
A no-tool ablation of GLM-5.1 solves $30.8\%$ [27.3,\,34.4] of the full core vs $26.6\%$ [23.4,\,30.1]
with tools---overlapping Wilson CIs, so the data do not support that tools help (I anchor to
the full core because on the frozen core the roster scores $0\%$ by construction). This is the
same-temperature counterpart of the agentic collapse of \S\ref{sec:exp:behavior}: GLM-5.1 is the model
most prone to abandoning its tools. The parity replicates on an independent, non-collapsing model:
GPT-5.5 solves $59.6\%$ [54.8,\,64.1] with tools vs $55.6\%$ [50.8,\,60.2] without. It is not a rubric
artifact: decomposing GLM-5.1's score by criterion type, the no-tool condition is at least as high on
every type---\texttt{must\_ground} included, exactly where retrieval should help
(Appendix~\ref{app:extended}).

\paragraph{Robust to the checklist judge and across domains.}
Two pre-specified checks confirm the range is not an artifact (detail in
Appendix~\ref{app:extended}): a different-family judge (Opus-4.7, $2{,}498$ pairs) shifts each solve
rate by $\le4.1$ points while preserving the ordering and the frontier-vs-held-out separation;
and across the $12$ L1 domains the frozen core spans all twelve, wrong-paper is comparable ($8$--$18\%$),
and the range holds within the largest---only agentic collapse is domain-dependent.

\begin{table}[t]
\centering
\small
\setlength{\tabcolsep}{3pt}
\resizebox{\columnwidth}{!}{%
\begin{tabular}{@{}llrr@{}}
\toprule
\textbf{Model} & \textbf{Developer} & \textbf{Full-core (657)} & \textbf{Frozen (423)} \\
               &                    & \textbf{solve\% [CI]} & \textbf{solve\% [CI]} \\
\midrule
\multicolumn{4}{@{}l}{\textit{Roster --- defines the buckets}}\\
GLM-5.1               & Zhipu    & $26.6$\,[23.4,\,30.1] & $0$\,\textsuperscript{$\ast$} \\
Qwen3.6               & Alibaba  & $11.7$\,[9.5,\,14.4]  & $0$\,\textsuperscript{$\ast$} \\
DeepSeek-V4           & DeepSeek & $6.2$\,[4.6,\,8.3]    & $0$\,\textsuperscript{$\ast$} \\
\midrule
\multicolumn{4}{@{}l}{\textit{Held-out --- same lineage as roster}}\\
GLM-5                 & Zhipu    & $26.1$\,[22.9,\,29.6] & $16.6$\,[13.3,\,20.4] \\
Qwen3.5-397B          & Alibaba  & $22.8$\,[19.7,\,26.1] & $16.8$\,[13.6,\,20.7] \\
Qwen3-235B (old)      & Alibaba  & $3.5$\,[2.3,\,5.2]    & $2.1$\,[1.1,\,4.0] \\
\midrule
\multicolumn{4}{@{}l}{\textit{Held-out --- independent lineage (closed-API frontier)}}\\
Gemini-3-Pro          & Google    & $37.4$\,[33.8,\,41.2] & $28.8$\,[24.7,\,33.3] \\
Opus-4.7              & Anthropic & $48.6$\,[44.8,\,52.4] & $37.8$\,[33.3,\,42.5] \\
GPT-5.5               & OpenAI    & $66.7$\,[63.0,\,70.2] & $59.6$\,[54.8,\,64.1] \\
\midrule
\multicolumn{4}{@{}l}{\textit{No-tool ablation}}\\
GLM-5.1 (no tools)    & Zhipu    & $30.8$\,[27.3,\,34.4] & $19.9$\,[16.3,\,23.9] \\
GPT-5.5 (no tools)    & OpenAI   & $60.8$\,[57.0,\,64.5] & $55.6$\,[50.8,\,60.2] \\
\bottomrule
\end{tabular}}
\caption{\textbf{Agentic leaderboard on the core set} ($T{=}0$, ten-tool harness;
no-tool ablations excepted); brackets are Wilson $95\%$ binomial CIs.
\textsuperscript{$\ast$}Roster frozen-core solve rate is $0$ by construction, so
\textbf{the two columns use different denominators and are not comparable}: a
frontier agent's frozen-core score is its rate on the subset hardest \emph{for the
roster}. Independent-lineage frontier agents span a wide capability range yet even
the best leaves a third or more unsolved (non-saturating); tool access confers no
measurable advantage. Full-core frontier cells combine frozen-core judging ($423$)
with a fresh $T{=}0$ pass on the $234$-item complement (\S\ref{sec:exp:repro}).}
\label{tab:core}
\vspace{-0.2cm}
\end{table}

\subsection{OpenBioRQ Measures What Closed-Form Medical QA Cannot}
\label{sec:exp:ortho}
\begin{figure*}[t]
\centering
\includegraphics[width=\linewidth]{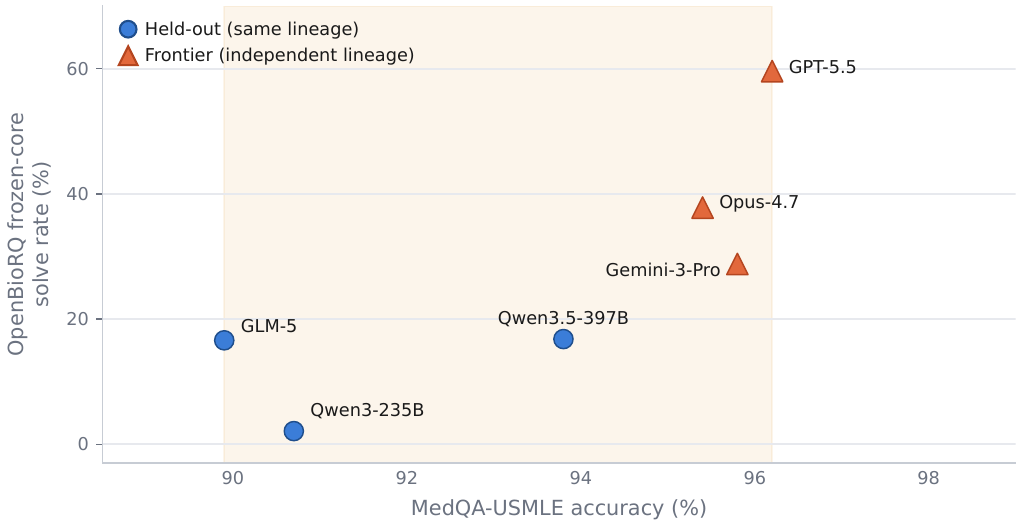}
\caption{Closed-form MedQA saturates while OpenBioRQ spreads. Held-out (blue, same-lineage) and
frontier (red, independent-lineage) models all fall in a narrow MedQA band (shaded), yet their
OpenBioRQ frozen-core solve rate ranges widely (the roster scores $0$ on the frozen core by
construction and is omitted; see Table~\ref{tab:ortho} for the full core). MedQA has saturated out of
discriminating range---a few points of MedQA map to a $\sim$30-point OpenBioRQ swing. This is a
resolution gap (MedQA cannot separate models OpenBioRQ separates), not a claim that the two
measure unrelated abilities; the weak rank correlations are reported in \S\ref{sec:exp:ortho}.}
\label{fig:ortho}
\end{figure*}

On MedQA the same six open-weight models sit in a saturated $3.9$-point band ($89.9$--$93.8\%$); on
OpenBioRQ they span $7.6\times$ (Figure~\ref{fig:ortho}, Table~\ref{tab:ortho}). The
dissociation is sharp pairwise---DeepSeek-V4 and GLM-5 are within $0.2$ MedQA points yet
$\sim$4$\times$ apart on OpenBioRQ, and the top MedQA model (Qwen3.5-397B) is beaten on
OpenBioRQ by GLM-5.1. This is a resolution gap, not zero shared signal: it is carried by the
band and pairwise inversions, not the weak rank correlations ($\rho{=}0.14$ core, $0.73$ three-exam
average; both $n{=}6$).

\begin{table}[t]
\centering
\small
\setlength{\tabcolsep}{3pt}
\resizebox{\columnwidth}{!}{%
\begin{tabular}{@{}lcccc@{}}
\toprule
 & \multicolumn{3}{c}{\textbf{Closed-form acc.\ (\%)}} & \textbf{OpenBioRQ} \\
\cmidrule(lr){2-4}
\textbf{Model} & \textbf{MedQA} & \textbf{PubMedQA} & \textbf{MedMCQA} & \textbf{core} \\
\midrule
GLM-5.1        & $91.0$ & $79.7$ & $79.9$ & $26.6$ \\
Qwen3.6        & $90.3$ & $80.5$ & $75.8$ & $11.7$ \\
DeepSeek-V4    & $90.1$ & $79.3$ & $78.9$ & $6.2$ \\
GLM-5          & $89.9$ & $80.0$ & $78.7$ & $26.1$ \\
Qwen3.5-397B   & $\mathbf{93.8}$ & $80.4$ & $80.6$ & $22.8$ \\
Qwen3-235B     & $90.7$ & $78.3$ & $77.5$ & $3.5$ \\
\midrule
range          & \multicolumn{3}{c}{MedQA $89.9$--$93.8$ ($3.9$\,pt)} & $3.5$--$26.6$ \\
\bottomrule
\end{tabular}}
\caption{OpenBioRQ measures a dimension closed-form medical QA does not: the six open-weight models
cluster in a narrow MedQA band (saturation) but spread widely on OpenBioRQ full-core solve rate, and
the best MedQA model is not the best on OpenBioRQ. Closed-form sets use deterministic decoding,
exact-match grading, $<\!0.5\%$ unparsed.}
\label{tab:ortho}
\end{table}

\subsection{Reproducibility of the Frozen Core}
\label{sec:exp:repro}
A second independent $T{=}0$ re-decode overturns my own difficulty headline: membership churns under
re-sampling ($46.5\%$ of boundary items and $34.7\%$ of near-threshold deep-failures flip), so the
frozen core is a single-sample snapshot. \textbf{I therefore retract an earlier $85.8\%$ retention
estimate and report no single retention figure.} The churn is a property of the selection, not
the results: every model is scored on the identical $423$-item set (a fixed released artifact), the
frontier range is unchanged on the churn-stable subset ($24.6/32.7/55.5\%$ on the $346$ items
stable across both decodes), and the citation and openness audits are computed elsewhere and
untouched. The core
is also not idiosyncratic to the three roster models: re-deriving the all-three-fail set from
alternate triples preserves Jaccard $0.70$--$0.77$ under a single swap and recovers $72\%$ under a full
same-lineage swap, so it is largely a property of the open-weight capability tier (full churn
and roster-robustness accounting in Appendix~\ref{app:extended}).

\section{Discussion and Limitations}
\label{sec:discussion}

\paragraph{Openness is dated, not eternal.}
Openness is an as-of property of \openbiorq{}, not a permanent one: a question open when its
source is written may resolve months later as a trial reads out. I pin it down with a
retrieval-grounded status verifier rather than parametric belief---a re-audit of the $657$ core found
\textbf{none} resolved (\S\ref{sec:benchmark})---and, because the property is dated, I release
per-item source records for independent re-verification and treat periodic re-auditing as a maintenance
obligation rather than a one-time guarantee.

\paragraph{Limitations.}
The robustness checks I lean on (cross-family judge swaps, per-domain breakdowns, re-decode
stability, roster-composition variation) are pre-specified probes of fixed headline claims, not a
search over hypotheses---and I report all of them, including the deep-failure test that
overturned my own $85.8\%$ retention figure.
\begin{itemize}[leftmargin=1.4em,itemsep=2pt,topsep=2pt]
  \item \textbf{The L2 wrong-paper rate is a judge-relative estimate; only a preliminary
  non-expert human spot-check, not domain-expert validation, exists so far.} An independent
  different-family judge and a full-text re-judge both leave the rate intact; the domain-expert
  $\kappa$ study is deferred (\S\ref{sec:experiments}; Appendix~\ref{app:discussion}).
  \item \textbf{Difficulty is a single $T{=}0$ snapshot and the frozen-core selection churns.} A
  second $T{=}0$ decode churns membership, so \textbf{I report no single retention figure}, though the
  range is essentially unchanged on items surviving both decodes (\S\ref{sec:exp:repro}). Difficulty
  is also model-relative---``hard'' means hard for today's open-weight roster, with the strongest
  frontier agent still solving a majority.
  \item \textbf{Scoring a new model still needs live tools.} Because re-running the live selection
  yields a substantially different core, I release the fixed $423$-item \texttt{task\_id} list and a
  tool-response replay cache rather than a re-runnable procedure; the full release mechanics are in
  Appendix~\ref{app:discussion}.
\end{itemize}

Extended discussion (construction-method transfer, the gold-citation sub-finding, licensing,
dual-use, and ethics) is in Appendix~\ref{app:discussion}.

\section{Conclusion}
\label{sec:conclusion}

I presented \openbiorq{}, a retrieval-grounded benchmark of unsolved biomedical research
questions for tool-using agents---the first agentic biomedical benchmark I am aware of to target
questions with no answer key, scoring grounding and abstention instead of a settled answer. Three
properties make it work, each answering a worry the open setting raises: openness is
verified rather than assumed (a retrieval-grounded status verifier corrects the confirmation
bias of source-framing refinement; narrative resolution remains undetectable by this audit),
difficulty is empirical rather than hand-labeled (a non-saturating frontier range on a
frozen core; \S\ref{sec:exp:core}), and evaluation is agentic by construction (multi-round
tool use over ten biomedical APIs graded by frozen checklist rubrics).

My central finding is that \textbf{existence is not correctness}: biomedical agents almost never
invent identifiers ($\approx 0.7\%$ over $4{,}863$ citations, the opposite of other
domains~\cite{researchmath14k}), yet roughly $15.9\%$ of their real citations are wrong-paper
($10.6\%$ under a different-family judge; a judge-relative estimate pending expert
validation)---a resolvable PMID attached to a claim it does not support, invisible to an L1-only
check. The same setting surfaces agentic collapse: on the hardest questions GLM-5.1 and
DeepSeek-V4 give zero-tool answers $65\%$/$62\%$ of the time, and a no-tool ablation confers no
measurable advantage. The takeaway for biomedical RAG and literature-synthesis systems is
concrete---an existence check is false comfort, so \textbf{L2 content support, not L1 existence, is
the reliability guardrail}, a caution that extends to LLM-generated reference answers
(\S\ref{sec:evaluation}). I will validate the checklist judge against human experts for a calibrated
Cohen's $\kappa$ and release the deferred SFT corpus from the $8{,}931$ trajectories already
collected. I release \openbiorq{} with per-item source records as a research-assistance instrument for
measuring and improving literature grounding---explicitly \textbf{not} clinical decision support, and
no score should be read as fitness to inform patient care.

\bibliographystyle{aaai}
\bibliography{references}

\appendix
\section*{Technical Appendix}

\section{Construction, Openness, and Difficulty Details}
\label{app:construction}

\paragraph{Construction pipeline.}
The corpus is built by a five-stage pipeline,
\textsf{crawl}~$\rightarrow$~\textsf{extract}~$\rightarrow$~\textsf{refine}~$\rightarrow$~\textsf{dedup}~$\rightarrow$~\textsf{export}.
In brief: authoritative open-problem sources are crawled; an LLM
extract agent lifts the distinct open questions each source raises (uncapped, so one paper can
yield several, each with a unique \texttt{task\_id}); an LLM refine agent rewrites every question
to stand alone and tags it with taxonomy, tool, and difficulty hints (self-containment
$51.6\%\!\rightarrow\!85.4\%$); near-duplicate question text is removed (MiniLM-L6 cosine
$\geq 0.90$); and the OpenBioRQ corpus of $12{,}553$ questions over $12$ domains is exported.
Two labelings are then layered on additively: retrieval-grounded openness (a status verifier feeding a
Stage-2 judge) and empirical, model-derived difficulty
(all-three-fail~$\rightarrow$~core~$\rightarrow$~frozen core at $T{=}0$).

\paragraph{Crawl.}
Documents are harvested from authoritative sources that frame open
problems rather than report closed results (the per-track sources are detailed in
\S\ref{sec:tracks}).

\paragraph{Extract.}
Each crawled document is passed to a Claude command-line extractor that lifts the
distinct open research questions it raises. Extraction is not capped at
one question per document: a single source paper can yield several questions.

\paragraph{Refine.}
A second Claude pass enriches each extracted question with (i) a taxonomy label
over $12$ level-one categories, (ii) three-axis
difficulty hints, (iii) a mapping to the relevant medical MCP tools, and (iv) an
initial \texttt{open\_status} with reasoning. The single-pass, tool-less status
assigned here is later replaced by a retrieval-grounded judgment.

\paragraph{Dedup.}
Near-duplicate questions are collapsed with a sentence-embedding screen: questions
whose MiniLM-L6 \cite{minilm} embeddings have cosine similarity
$\geq 0.90$ are treated as duplicates. The dedup is over question text, so
the multiple distinct questions extracted from one source paper are retained;
only genuinely identical questions collapse.

\paragraph{Export.}
The deduplicated questions are exported as the OpenBioRQ corpus, organized into four named
tracks---\texttt{retrieval\allowbreak\_verified}, \texttt{expert\_consensus}, \texttt{priority\_setting},
and \texttt{expand}---whose sizes and intended uses are given under Corpus statistics below.

\subsection{Self-Containment Audit}
A question is only useful in isolation if it can be read without its source document; the refine pass
is what makes it standalone. On an LLM-judged random sample of the refined output ($n=500$),
$85.4\%$ of questions are self-contained, up from $51.6\%$ for the extractor-only baseline---a
$+33.8$-point lift attributable to the refine pass alone (Table~\ref{tab:construction}). The gain is
uneven across tracks: the \texttt{retrieval\allowbreak\_verified} track comes out substantially
cleaner ($95\%$) than \texttt{expert\_consensus} ($75\%$).

\begin{table}[t]
\centering
\small
\setlength{\tabcolsep}{3pt}
\resizebox{\columnwidth}{!}{%
\begin{tabular}{@{}lr@{}}
\toprule
\textbf{Construction-quality metric} & \textbf{Value} \\
\midrule
Self-containment, overall ($n{=}500$)            & $85.4\%$ \\
\quad retrieval\_verified track                  & $95\%$ \\
\quad expert\_consensus track                    & $75\%$ \\
Self-containment, extractor-only                 & $51.6\%$ \\
\quad refine pass contribution                    & $+33.8$ pts \\
\midrule
Dedup threshold (MiniLM-L6 cosine)               & $\geq 0.90$ \\
\midrule
Status flip (source$\to$retrieval, $n{=}200$)    & $56.5\%$ \\
Still-open audit, core set                       & $0/657$ resolved \\
\midrule
Hard contamination, core set ($n{=}657$)         & $0\%$ \\
Hard contamination, expand track ($n{=}483$)     & $0\%$ \\
\bottomrule
\end{tabular}}
\caption{Construction-quality summary: self-containment lift from the refine pass, dedup threshold,
grounded status-flip rate, and the still-open / hard-contamination audits.}
\label{tab:construction}
\end{table}

\subsection{Retrieval-Grounded Openness}
A question's openness is judged from real follow-up evidence, not from a model's
memory of how the source framed the problem. The original source-framing-only
refiner produced complete confirmation bias: the corpus contained zero
\texttt{answered} or \texttt{unknown} labels. A retrieval-grounded Stage-2 judge
re-decides openness only from gathered follow-up evidence, making resolution
detectable.

\paragraph{Stage 1: evidence gathering.}
A non-LLM verifier (\texttt{status\allowbreak\_verifier.py}) dispatches on the
\texttt{source\_id} type to gather follow-up evidence: PubMed PMIDs are resolved
to their citing papers and abstracts via the Europe PMC citation index
\cite{europepmc}; trial NCT identifiers are checked against ClinicalTrials.gov v2
status, completion, and results \cite{clinicaltrials}; and arXiv identifiers are
expanded through Semantic Scholar citations \cite{semanticscholar}. (I found
NCBI \texttt{elink} ``citedin'' unreliable for PMIDs and use Europe PMC as the
primary follow-up source.) The bundle of gathered evidence IDs and snippets is
attached to each question.

\paragraph{Stage 2: grounded re-judgment.}
A Claude judge (\texttt{stage2\_judge.py}) re-decides \texttt{open\_status}
only from the gathered evidence and must cite evidence IDs present in the
bundle; hallucinated citations are rejected and the item falls back to
\texttt{unknown}, so the \texttt{answered}/\texttt{unknown} labels that source-framing
never reached become attainable.
On a random sample of 200 retrieval-verified questions, grounded re-judgment
changes the status label of $56.5\%$ of them and assigns \texttt{unknown} to $14\%$
(unreachable under source-framing), confirming that the ``open'' label is grounded in
evidence rather than an artifact of the source's framing.

\paragraph{Still-open audit.}
To confirm that the hardest questions genuinely survive grounded re-judgment, I
ran a still-open audit over the $657$ core questions (the trial-completed and
no-follow-up flagged subset). Under grounded re-judgment, $0/657$ were confirmed
resolved---no detectable resolution. Narrative or
guideline resolution is undetectable by this audit, and its recall is unquantified.
A positive control validates the structured-contamination detector: on $34$ injected
known-contaminated items (real completed-with-results trials relabeled open, and templated source
ids) it achieves $100\%$ recall ($34/34$) and $0/12$ false-positives on genuine-open negatives---not
a blind detector.
I do not claim the same for the openness verifier's recall
against narrative literature resolution (a confirmatory study or guideline that is not a
trial-status field or template): that is bounded by my conservative, evidence-based criterion
rather than positive-controlled, and may miss not-yet-indexed consolidation.

\paragraph{Membership / anti-memorization check.}
As a further membership check, the evaluated question string is the LLM-refined
\texttt{self\_contained\allowbreak\_question}, not source text. Over the frozen core it shares a
median longest-verbatim span of only $3$ words with its source phrasing (median question length $22$
words; token Jaccard $0.23$), and $86.5\%$ of questions share $\le 5$ consecutive verbatim words. So
exact-string memorization of a benchmark question is largely precluded by construction---this
verbatim-overlap result is the load-bearing membership argument.

A model-based check is consistent with it: under GLM-5.1 the frozen-core questions are
higher-perplexity than MedQA-USMLE---a long-public benchmark very likely in pretraining
corpora---by $1.61$ nats/token (mean log-prob $-3.22$ vs $-1.61$, $n{=}200$ each), i.e.\ the core
shows no stronger memorization signature than a set I already expect to be seen. I treat
this as corroborating rather than independent: per-token perplexity confounds memorization with
question length and syntactic complexity, and the probe uses a single roster-lineage model, so it
should not be over-read.

Both checks target surface memorization (exact strings, perplexity) and would not detect
paraphrased exposure of a question's underlying problem in pretraining; since my items are
LLM-rephrasings of public sources this cannot be excluded---though with no answer key to recall, such
exposure confers little advantage, since rote retrieval of a settled answer is exactly what an open
question denies.

\subsection{Empirical Difficulty}
Difficulty is read off model behavior rather than assigned by hand. Every question is answered with
full tool access by the three roster models---GLM-5.1~\cite{glm}, Qwen3.6~\cite{qwen}, and
DeepSeek-V4~\cite{deepseek}---and graded against its frozen checklist rubric. A question is
labeled core exactly when all three fail it (checklist score $<0.5$). On the cleanest,
question-granular \texttt{priority\_setting} track ($n{=}525$), $49\%$ of questions are core, $45\%$
discriminating, and $6\%$ easy ($256/236/33$ of $525$; the bucket composition and per-model mean
scores are Figure~\ref{fig:difficulty} in the main body).

\paragraph{Rejected source: preprints.}
I trialed medRxiv \cite{medrxiv} abstracts as an additional source and rejected
them. Preprint abstracts overwhelmingly report results rather than frame
open questions, so extraction yields almost nothing usable; crawling is instead
concentrated on priority-setting and research-gap documents.

\paragraph{Question granularity.}
Because extraction produces several distinct questions per source paper, the unit
of evaluation must be the question, not the paper. Every question therefore
carries a unique \texttt{task\_id} of the form \texttt{source\_id\#k} (e.g.,
\texttt{PMID:38345416\#0}). Keying rubrics, traces, and judgments on
\texttt{source\_id} alone would silently collapse co-extracted questions to
source-paper granularity and conflate their (often different) difficulties; the
\texttt{\#k} suffix prevents this collapse, keeping co-extracted questions as the distinct items they
are.

\paragraph{Corpus statistics.}
\openbiorq{} is a single dataset organized into four named tracks. The
\texttt{retrieval\allowbreak\_verified} ($6{,}648$) and \texttt{expert\_consensus}
($5{,}905$) tracks form the $12{,}553$-question base corpus; the
\texttt{priority\_setting} track adds $525$ questions and the \texttt{expand} track
adds $483$, each deduplicated against the base. A gold-answer-bearing slice of
$1{,}969$ questions (\texttt{mcp\_benchmark\allowbreak\_with\_gold.jsonl}) is used for rubric
generation and agentic evaluation. Openness is an as-of property: openness
labels and difficulty scores reflect their snapshot date and are reproducible against
that snapshot. The live literature and tool APIs move, however, so labels gathered at
different dates may differ; re-running the affected audits refreshes them. The corpus distribution
across the $12$ level-one taxonomy categories is shown in the main body (Figure~\ref{fig:taxonomy}).

\paragraph{Example questions and checklists.}
To make the unit of evaluation concrete, I show two representative items verbatim, each with its
frozen per-question checklist of weighted criteria typed \texttt{must\_mention} /
\texttt{must\_acknowledge} / \texttt{must\_ground} / \texttt{must\_avoid} (\S\ref{sec:evaluation}).

\smallskip
\noindent Example 1 --- Pharmacology \& Drug Discovery, \texttt{open}.
``Can therapeutic interventions targeting the glymphatic system (the brain's perivascular waste
clearance pathway) be used to prevent or slow the progression of Alzheimer's disease?''
\begin{itemize}[leftmargin=1.4em,itemsep=1pt,topsep=2pt]
  \item \texttt{must\_mention} (w3): sleep enhancement or orexin-receptor antagonists (e.g.\ suvorexant) as preclinical interventions that enhance clearance.
  \item \texttt{must\_mention} (w2): AQP4 polarization/localization as a key mechanistic target.
  \item \texttt{must\_acknowledge} (w3): no therapy specifically targeting the glymphatic system has entered clinical trials for AD.
  \item \texttt{must\_acknowledge} (w3): enhancing glymphatic clearance in humans may not be sufficient to meaningfully alter AD.
  \item \texttt{must\_ground} (w2): cites real preclinical evidence (e.g.\ Xie et al.\ 2013, Lundgaard et al.).
  \item \texttt{must\_avoid} (w3): does not claim glymphatic-targeting therapies are proven to prevent or slow AD in humans.
  \item \texttt{must\_avoid} (w2): does not fabricate clinical-trial results or citations for human efficacy.
\end{itemize}

\smallskip
\noindent Example 2 --- Medical AI \& Informatics, \texttt{open}.
``How should astrocytes be incorporated into mechanistic, theoretical, and computational models of
neural circuits to better understand the pathophysiology of neurological and psychiatric diseases
with known astrocytic involvement?''
\begin{itemize}[leftmargin=1.4em,itemsep=1pt,topsep=2pt]
  \item \texttt{must\_mention} (w3): the tripartite-synapse model and calcium-dependent gliotransmitter release.
  \item \texttt{must\_mention} (w2): mean-field or network models incorporating astrocytic glutamate uptake or potassium buffering.
  \item \texttt{must\_acknowledge} (w3): no widely adopted, standardized computational framework currently exists for integrating astrocytes.
  \item \texttt{must\_acknowledge} (w3): the difficulty of defining astrocytic computational primitives (e.g.\ spatially distributed calcium signals).
  \item \texttt{must\_ground} (w2): cites real established frameworks (e.g.\ Araque et al., De Pitta et al.).
  \item \texttt{must\_avoid} (w3): does not claim a universally accepted astrocyte framework already exists.
  \item \texttt{must\_avoid} (w2): does not fabricate citations or falsely claim tools returned nothing.
\end{itemize}
The \texttt{must\_avoid} criteria operationalize the wrong-paper and fabrication hazards directly at
grading time, and the \texttt{must\_acknowledge} criteria reward calibrated abstention on what
remains unsettled.

\section{Definitions, Notation, and Harness}
\label{app:formalism}

\begin{figure*}[t]
\centering
\includegraphics[width=\linewidth]{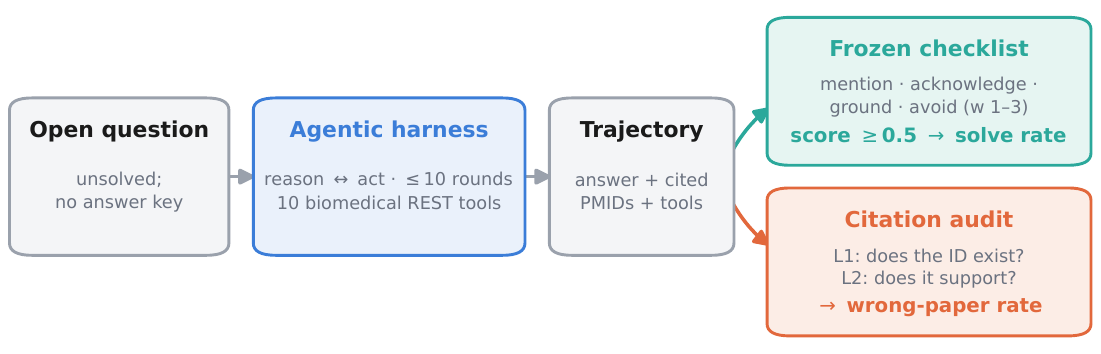}
\caption{The \openbiorq{} evaluation flow (companion to the construction pipeline in
Appendix~\ref{app:construction}). One open question---with no answer key, plus its frozen
per-question checklist---is answered by an agentic harness that interleaves reasoning and tool calls
(up to ten rounds) over ten biomedical REST APIs. The resulting trace (final answer, cited
PMIDs, tool calls) is graded by two independent measures: the frozen checklist (weighted;
$\ge0.5$ = solve) yields the \textbf{solve rate}, and the two-level citation audit (L1 existence,
L2 support) yields the \textbf{wrong-paper rate}---the apparatus behind my difficulty and citation
findings, respectively.}
\label{fig:evalflow}
\end{figure*}

This appendix makes the evaluation of \S\ref{sec:evaluation} precise, one definition at a time.
Figure~\ref{fig:evalflow} sketches the end-to-end flow the definitions below formalize. I
start from the per-question checklist score and the solve rate it thresholds; build the
empirical-difficulty buckets that carve out the core and frozen core; and state the
citation and openness predicates behind the wrong-paper and grounded-openness numbers. I close with
why a frozen checklist, rather than a free-form judge, is what keeps these definitions stable when an
agent collapses.

\paragraph{Checklist score.}
The score of an answer is how much of its frozen checklist it satisfies, weighted by each
criterion's importance. In detail: for a question $q$ with frozen checklist
$C_q=\{(t_i,w_i)\}_{i\in I_q}$, each criterion has a type $t_i$---one of \texttt{must\_mention},
\texttt{must\_acknowledge}, \texttt{must\_ground}, \texttt{must\_avoid}---and a weight
$w_i\in\{1,2,3\}$, with $|I_q|\in[5,8]$. A judge assigns a verdict $v_i\in\{1,0.5,0\}$ via the map
$\mu:\{\text{met},\text{partial},\text{not\_met}\}\to\{1,0.5,0\}$; the partial value $0.5$ is the
midpoint of the verdict scale, chosen for interpretability rather than fit. I make no robustness
claim to its exact value. One type runs the other way. For \texttt{must\_avoid} criteria the polarity
is inverted: $v_i{=}1$ means the prohibited behavior---fabricating a citation, or asserting a
definitive answer to an open question---was correctly avoided, so a higher score never rewards
confabulation. The question score is then
$\mathrm{score}(q)=\big(\sum_{i\in I_q} w_i v_i\big)/\big(\sum_{i\in I_q} w_i\big)$
(Eq.~\ref{eq:checklist}), and it is well-defined and bounded: because $w_i\ge1$ and
$|I_q|\ge5$ the denominator $\sum_{i\in I_q} w_i>0$, and since each $v_i\in[0,1]$ the score lies in
$[0,1]$---a range the \texttt{must\_avoid} inversion preserves.

\paragraph{Solve rate.}
The \textbf{solve rate}, my headline metric, is the plainest quantity the score supports: the
fraction of a fixed question set a model solves under a single deterministic try. Precisely, a model
$m$ \emph{solves} $q$ iff $\mathrm{score}^T_m(q)\ge\tau$ with $\tau{=}0.5$ (the superscript $T$ records
the decoding temperature), and the headline number fixes $T{=}0$---deterministic, one attempt. I
avoid the label ``pass$@k$'' on purpose: there $k$ is a \emph{sample count}, whereas my $0.5$ is a
score \emph{threshold} over a fixed denominator. Symmetrically, $m$ \emph{fails} $q$ iff
$\mathrm{score}^T_m(q)<0.5$, and the difficulty threshold used throughout the main text is exactly this
$\tau{=}0.5$ instance.

\paragraph{Empirical difficulty, core, and frozen core.}
Difficulty here is not assigned by hand; it is read off the roster. Fix the roster $M$ of
models---$|M|{=}3$, namely $\{$GLM-5.1, Qwen3.6, DeepSeek-V4$\}$---and let
$F_T(q)=|\{m\in M:\mathrm{score}^T_m(q)<0.5\}|$ count how many of them fail $q$ at temperature $T$. The
empirical-difficulty bucket is just that count: $q$ is \texttt{all3\_fail} iff $F_{0.3}(q)=|M|$,
\texttt{all3\_pass} iff $F_{0.3}(q)=0$, and \texttt{split} otherwise. The \textbf{core} keeps only the
all-fail items that also survive the audits, as below,
\begin{multline*}
  \mathcal{C}=\{q: F_{0.3}(q)=|M| \;\wedge\; \mathrm{resolved}(q){=}0 \\
  {}\wedge\; \mathrm{hardContam}(q){=}0\},
\end{multline*}
i.e.\ the all-roster-fail items---at the temperature $T{=}0.3$ at which the buckets were originally
defined---that additionally pass the openness and contamination checks, where
$\mathrm{resolved}(q),\mathrm{hardContam}(q)\in\{0,1\}$ are the indicator outcomes of the audits of
\S\ref{sec:grounded} (detailed in Appendix~\ref{app:construction}). The \textbf{frozen core}
then tightens this one step further, keeping only the items that stay all-roster-fail under
deterministic decoding,
\begin{equation*}
  \mathcal{C}^\star=\{q\in\mathcal{C}: F_{0}(q)=|M|\}\subseteq\mathcal{C},
\end{equation*}
which removes items that only looked hard under stochastic decoding---so $423\subseteq657$ by
construction.

\paragraph{Citation audit and openness predicates.}
Two predicates turn raw audit outcomes into the numbers I report. Take citations first. Each emitted
identifier carries a fetch outcome
$\mathtt{fetch}\in\{\mathtt{found},\mathtt{notfound},\mathtt{transient}\}$, and \textbf{L1} existence
counts $\mathtt{notfound}$ as fabrication while excluding $\mathtt{transient}$. For a $\mathtt{found}$
citation, an \textbf{L2} support label $\mathtt{supp}\in\{\mathtt{yes},\mathtt{partial},\mathtt{no}\}$
asks whether the cited record actually substantiates the claim; the \textbf{wrong-paper} case is the
narrow one, $\mathtt{found}\wedge\mathtt{supp}{=}\mathtt{no}$, and I compute wrong-paper rates over
$\mathtt{found}$ citations only. Openness is the second predicate. The Stage-2 status $S(q)$ takes
values in $\{\mathtt{open},\mathtt{partially\_answered},\mathtt{answered},\mathtt{unknown}\}$ and is
tied to the gathered evidence set $E(q)$: every identifier it cites must lie in that set,
$\mathrm{cites}(S(q))\subseteq E(q)$ (\S\ref{sec:grounded}). Agreement is always
measured between judge pairs---Spearman over per-answer checklist scores for the LLM-vs-LLM
correlation, and Cohen's $\kappa$ over the wrong-paper categories.

\paragraph{Rubric generation.}
The rubric generator is held to one rule: every criterion must be binary-checkable and
instance-grounded. ``Identifies AQP4 depolarization as a candidate mechanism'' is admissible;
``Discusses the mechanism well'' is not. It must also emit at least one \texttt{must\_acknowledge} and
one \texttt{must\_avoid} criterion per question, so that calibrated uncertainty and the absence of
fabrication are always scored rather than left to the judge's discretion.

\paragraph{Robustness to agentic collapse.}
A practical advantage of checklist scoring is its robustness to agentic collapse. On the
hardest questions, models frequently degrade into non-attempts or zero-tool traces. A continuous
free-form judge tends to award such empty or hedging answers partial credit for fluent prose---blurring
the very distinction the benchmark is designed to expose. The frozen checklist instead
withholds credit precisely where it matters: a collapsed answer fails its \texttt{must\_mention} and
\texttt{must\_ground} criteria and, if it falsely claims the tools yielded nothing, also fails
\texttt{must\_avoid}, so the score reflects the absence of grounded substance rather than surface
fluency. This keeps the difficulty buckets stable and makes the resulting core set a reliable target
for measuring progress.

\section{Extended Experiments and Robustness}
\label{app:extended}

Each headline in \S\ref{sec:experiments} is a pooled number; here I open it up. I begin with the
citation audit at full per-model resolution and its robustness probes; the behavioral breakdown; the
no-tool ablation by criterion; two robustness checks (checklist judge, domains); and the
reproducibility accounting behind the frozen core.

\subsection{Citation Audit: Robustness Probes}
The full per-model L1/L2 breakdown is Figure~\ref{fig:citation} in the main body; here I report the
robustness probes behind the headline wrong-paper rate.

\paragraph{Conservative floor: the partial bin.}
Wrong-paper excludes the large ``partial'' bin ($28.1\%$ of evaluable citations under the Opus
judge). A strict binary re-judge of substantive partials (tangential, wrong-population, or
wrong-direction support counted as not-supporting) flips a sizable fraction to ``no'' (in a
$25$-item probe, $10/25 = 40\%$ flip), so folding partials in would push the upper bound to
$\sim$$25\%$; the ``no''-only definition under-reports rather than inflates the failure.

\paragraph{Wrong-paper by identifier type.}
Splitting by identifier type, trial-registry citations are misused more often than publication
citations: under the Opus judge, NCT trial IDs are wrong-paper $20.3\%$ ($47/231$) versus $13.0\%$
for PMIDs ($593/4{,}545$)---mirroring their higher fabrication at L1 (GLM-5.1's invented identifiers
were predominantly NCT IDs).

\paragraph{Wrong-paper is a sourcing defect, not a marker of ungrounded reasoning.}
Across $879$ jointly-labeled answers (those with audited citations), a wrong-paper citation shows no
detectable association with whether the answer satisfies the \texttt{must\_ground} criterion: risk
ratio $1.07$ ($95\%$ CI $0.88$--$1.31$). The decoupling is not an artifact of a lenient grounding
check. Of the \texttt{must\_ground}-passing answers that carry a wrong-paper citation, $73\%$ also
carry a genuinely supporting one (every answer cites $\geq 2$ sources); and under a strict grounding
definition (requiring an actually-supporting citation) the association is still null (RR $1.06$,
$95\%$ CI $0.99$--$1.13$; odds-ratio point estimate $1.33$ [$0.93$,\,$1.89$], so independence is
not refuted rather than positively established). The join is at answer rather than per-claim
granularity. The takeaway: wrong-paper is a sourcing-integrity defect that persists even in answers
that otherwise look well-grounded, so neither an existence check nor a grounding-reward signal would
surface it.

\subsection{Behavioral Divergence}
The starkest gap is the cite-rate: it spans roughly $10\times$ across the roster---GLM $3.9\%$ versus
DeepSeek-V4 $38.5\%$---so the three models barely agree on whether to ground at all---and that is only
one axis; they diverge as sharply on the others. The full per-model breakdown over the
$1{,}969$-question set is in Table~\ref{tab:behavior} (the figure is in the main body,
Figure~\ref{fig:behavioral}).

\begin{table}[t]
\centering
\small
\begin{tabular}{@{}lrrrr@{}}
\toprule
\textbf{Model} & \textbf{Non-} & \textbf{Zero-tool} & \textbf{Avg} & \textbf{Cite-} \\
               & \textbf{attempt} & \textbf{collapse} & \textbf{tools} & \textbf{rate} \\
\midrule
GLM-5.1      & $26.2\%$ & $20.8\%$ & $12.6$ & $3.9\%$ \\
Qwen3.6      & $19.5\%$ & $19.6\%$ & $6.1$  & $23.9\%$ \\
DeepSeek-V4  & $0.8\%$  & $31.3\%$ & $8.3$  & $38.5\%$ \\
\bottomrule
\end{tabular}
\caption{Behavioral analysis on the 1{,}969-question set. Models diverge sharply on
every axis. Cite-rate spans roughly $10\times$ (GLM $3.9\%$ vs.\ DeepSeek-V4 $38.5\%$).}
\label{tab:behavior}
\end{table}

\paragraph{Agentic collapse is concentrated on the hardest track---and multi-model.}
On the broad $1{,}969$-set the zero-tool rate is only moderate for all three roster models
(Table~\ref{tab:behavior}: $20.8$/$19.6$/$31.3\%$). Collapse is specific to the harder,
question-granular priority track, where two of the three jump sharply: GLM-5.1 to $65.3\%$
zero-tool (non-attempt $69.1\%$, up from $26.2\%$) and DeepSeek-V4 to $62.3\%$ (non-attempt
$62.3\%$, up from $0.8\%$), while Qwen3.6 barely moves ($22.1\%$ zero-tool,
non-attempt $26.3\%$). Because collapse spans two independent developers (Zhipu and DeepSeek) it is
not a single-model idiosyncrasy; because the third (Alibaba's Qwen3.6) resists, it is not forced by
the questions alone. It is a model-divergent, non-monotone failure, not a uniform law---a structural
axis of open-question behavior that answer-only benchmarks cannot register.

\subsection{No-Tool Ablation: Criterion Decomposition}
The tool-parity result is not a rubric artifact. Decomposing GLM-5.1's score by criterion type over
the $656$ matched core questions, the no-tool condition scores at least as high on every type,
and its largest gains are on \texttt{must\_mention} (factual recall, $32.5\!\to\!40.2$) and
\texttt{must\_avoid} ($77.7\!\to\!84.7$), not on the abstention-credit \texttt{must\_acknowledge}
($5.5\!\to\!7.5$, the smallest gain). Even on \texttt{must\_ground}---which rewards citing real
supporting evidence, exactly where retrieval should help---tools confer no advantage ($17.1$ with
tools vs $19.3$ without).

\subsection{Leaderboard Robustness to the Checklist Judge}
Because the leaderboard is graded by a GLM-5.1 checklist judge (same lineage as a roster model), I
re-grade the frozen-core solve rate of all six non-roster models with an independent
different-family judge (Opus-4.7, $2{,}498$ question-model pairs). Per-model solve rate shifts by at
most $4.1$ points (Gemini-3-Pro $28.8\!\to\!27.8$, Opus-4.7 $37.8\!\to\!40.5$, GPT-5.5
$59.6\!\to\!58.4$, held-out GLM-5 $16.6\!\to\!20.7$); the per-question solve label agrees at Cohen's
$\kappa=0.65$, and both the frontier ordering (Gemini $<$ Opus $<$ GPT-5.5) and the
frontier-vs-held-out separation (min frontier $27.8\%$ $>$ max held-out $20.7\%$) survive intact (the
Opus rates are over the successfully-parsed subset of the $423$ per model). The capability range
is therefore not an artifact of the GLM-lineage judge.

\subsection{Per-Domain Breakdown}
Because every headline is a pooled number, I re-aggregate each by the L1 taxonomy
(Table~\ref{tab:domain}) and ask which findings survive the split. Three hold up; one does not. Take
wrong-paper first: its Opus rate runs from $8.1\%$ in Surgical to $18.1\%$ in Genomics---a
narrow band in which no single specialty drives the pooled rate, so the defect is not a
Clinical-Medicine artifact. The frozen core is similarly spread: it spans all twelve domains,
the largest share being Clinical Medicine at $21.5\%$, so it is not a single-specialty benchmark wearing
a general label. The frontier capability range survives the split too, and holds
within domains: in each of the four largest the Gemini\,$<$\,Opus\,$<$\,GPT-5.5 ordering is
preserved and GPT-5.5 solves a majority---Clinical Medicine $24\!\to\!38\!\to\!62$, Neuroscience
$23\!\to\!37\!\to\!62$, Oncology $27\!\to\!38\!\to\!60$, Infectious Disease
$38\!\to\!48\!\to\!50$---with only minor inversions in low-$n$ domains. The one finding that does
not survive uniformly is agentic collapse: GLM-5.1's zero-tool rate peaks in Medical
AI \& Informatics ($43.9\%$) and Surgical Sciences ($29.0\%$) but falls to $10.5\%$ in Oncology,
suggesting collapse is triggered more where the biomedical APIs return sparse hits.

\begin{table}[t]
\centering
\small
\setlength{\tabcolsep}{3pt}
\begin{tabular}{@{}lrrrrrr@{}}
\toprule
 & \textbf{rc} & \textbf{wp} & \multicolumn{3}{c}{\textbf{frontier solve rate}} & \textbf{coll.} \\
\cmidrule(lr){4-6}
\textbf{Domain} & \textbf{\%} & \textbf{\%} & \textbf{Gem} & \textbf{Opus} & \textbf{GPT} & \textbf{\%} \\
\midrule
Clinical Med   & $21.5$ & $13.7$ & $24.2$ & $38.5$ & $61.5$ & $25.2$ \\
Neuro/Psych    & $19.9$ & $12.0$ & $22.6$ & $36.9$ & $61.9$ & $16.6$ \\
Oncology       & $13.0$ & $14.6$ & $27.3$ & $38.2$ & $60.0$ & $10.5$ \\
Infect/Immun   & $9.5$  & $15.2$ & $37.5$ & $47.5$ & $50.0$ & $22.2$ \\
Public Health  & $6.6$  & $12.6$ & $35.7$ & $25.0$ & $53.6$ & $23.8$ \\
Surgical       & $5.9$  & $8.1$  & $48.0$ & $40.0$ & $72.0$ & $29.0$ \\
Cardiovascular & $5.7$  & $13.3$ & $41.7$ & $37.5$ & $54.2$ & $14.9$ \\
Genomics       & $5.7$  & $18.1$ & $25.0$ & $25.0$ & $70.8$ & $24.8$ \\
Pharm/Drug     & $4.7$  & $14.1$ & $30.0$ & $50.0$ & $70.0$ & $20.7$ \\
Medical AI     & $4.3$  & $12.8$ & $11.1$ & $33.3$ & $38.9$ & $43.9$ \\
Rare Disease\textsuperscript{$\dag$}   & $2.1$ & $12.4$ & $44.4$ & $55.6$ & $44.4$ & $20.0$ \\
Other\textsuperscript{$\dag$}          & $1.2$ & $21.6$ & $20.0$ & $20.0$ & $60.0$ & $0.0$ \\
\bottomrule
\end{tabular}
\caption{Per-domain breakdown over the $12$ L1 taxonomy categories. \textbf{rc\%}: share of the
$423$ frozen core; \textbf{wp\%}: Opus-judge wrong-paper rate; \textbf{frontier solve rate}:
Gemini-3-Pro / Opus-4.7 / GPT-5.5 on the frozen core; \textbf{coll.\%}: GLM-5.1 zero-tool collapse
rate. The frontier range (Gem $<$ Opus $<$ GPT) holds in every domain with $n\!\ge\!18$, and
wrong-paper is broadly comparable across domains.
\textsuperscript{$\dag$}$n\!<\!10$ frozen-core items---not interpretable.}
\label{tab:domain}
\end{table}

\subsection{Reproducibility of the Frozen Core: Full Accounting}
I re-decode all three roster models a second time at $T{=}0$, five days later (testing live-API
and agentic nondeterminism; the model output itself is deterministic), and re-judge on the same rubric.
Per-question scores reproduce closely (mean absolute checklist-score delta $0.09$--$0.14$;
fail-label agreement $0.78$--$0.81$). Membership, however, churns: of the boundary-proximal items
$46.5\%$ flip, and a direct test of the near-threshold ``deep'' failures (max-of-3 in $[0.35,0.40)$)
finds $34.7\%$ ($17/49$) also flip to pass---driven by live-API drift and erratic agentic tool use
(Qwen3.6 and DeepSeek-V4 account for $19$ of $23$ flips). I therefore \textbf{retract an earlier
$85.8\%$ retention estimate} (which had assumed deep-failures fixed) and report no single retention
figure; the frozen core is a single-$T{=}0$ snapshot, not a seed-stable partition.

The churn is a property of the selection, not the comparisons. Recomputing the frontier
range on the $346$ items stable across both decodes gives $24.6/32.7/55.5\%$ (vs the full-$423$
$28.8/37.8/59.6\%$)---same ordering and $\sim$31-point spread---and the $77$ flipped items are simply
easier for the frontier agents ($48.1/61.0/77.9\%$). The core is also not idiosyncratic to the three
roster models: re-deriving the all-three-fail set from alternate triples preserves Jaccard
$0.70$--$0.77$ under a single-model substitution and recovers $72\%$ (Jaccard $0.58$) under a full
same-lineage roster swap, so the core is largely a property of the open-weight capability tier.
I test only removals; whether a re-decode adds items with a different frontier profile
would require re-decoding the non-core questions, which I leave to future work.

\section{Extended Discussion, Release, and Ethics}
\label{app:discussion}

\paragraph{Why citation severity is a medical, not a math, problem.}
My two-level audit is motivated by a severity asymmetry I hypothesize is specific to
medicine. In a domain
like mathematics, a fabricated reference is academic sloppiness~\cite{walters2023fabricated,researchmath14k};
in medicine, I argue, a real identifier (a resolvable PMID or NCT
number) attached to an unsupported clinical claim is plausibly a more consequential
misattribution, because the resolving identifier can lend false authority and a
downstream reader may be more likely to trust the claim because the link works.
(This is a sourcing and attribution argument, not a clinical-harm measurement: I do not study
patient impact, claim veracity, or reader trust.) This is why
L1-existence alone is the wrong instrument: emitted identifiers almost always exist
($\approx 0.7\%$ non-existent over 4{,}863 citations, the opposite of ResearchMath's
$\approx 54\%$ fabrication rate), so the meaningful failure mode lives at L2---real papers
cited for claims they do not support ($15.9\%$ of real citations overall, a judge-relative LLM
estimate---see the non-expert spot-check below). Existence is not correctness, and in biomedicine the
gap between them is the reliability story.

\paragraph{Open questions as a faithfulness-and-abstention probe.}
Hand a model an open question and it has two basic moves: \emph{abstain}---honestly flag
that the literature does not settle the matter---or \emph{confabulate}---assert a confident
answer, often pinned to an unsupporting citation. Both of my headline findings are failures of this
choice. Agentic collapse is abstention taken to its degenerate extreme:
the agent stops engaging altogether rather than commit. Wrong-paper citation is the opposite
failure, confabulation dressed in a resolvable source link: a definitive answer made to look
sourced. Closed-form QA can expose neither, because the answer is already given; open questions
are what make the abstain-vs-confabulate trade-off the object of measurement.

\paragraph{Empirical, model-derived difficulty as a transferable construction method.}
The difficulty-labeling recipe---answer every candidate with a roster of models and
bucket by all-model failure (solve rate)---is a contribution independent of the biomedical
content, and it earns that independence through three properties. It needs no hand-labeling:
difficulty is read off the roster's answers rather than assigned by an annotator, so a human
never has to judge what is hard. It scales with corpus size: adding candidates costs only more
model passes, not more human effort, so the labeled pool grows as cheaply as the raw pool. And
it is intrinsically non-saturating: as models improve, items that were ``core'' migrate to
``discriminating'' and a fresh core can be redefined against the current frontier, so the
benchmark never permanently bottoms out the way a fixed human-labeled set does. The method is
domain-agnostic---it should transfer to any setting where a frozen, checkable rubric can be
generated, a reusable way to build hard benchmarks without human difficulty annotation.

My own two corrections carry the generalizable lessons for anyone using this recipe in an
agentic setting. First, empirical difficulty measured under stochastic decoding can be
partly a sampling artifact: my ``solve rate $\approx0\%$'' core dissolved to $\sim$1 in 4 solvable
once I re-measured at $T{=}0$---so pipelines that bucket by model failure must fix and report the
decoding temperature.

Second, even at $T{=}0$ the signal is not
reproducible if the environment is live, because run-to-run tool drift and erratic agentic behavior
churn membership ($34.7\%$ of even my near-threshold ``deep'' failures flip on a re-decode,
\S\ref{sec:exp:repro}). The recipe for agentic difficulty is therefore not just ``bucket by
all-model failure'' but ``bucket by all-model failure over a frozen tool environment''---one
must record and replay the tool responses (a difficulty signal taken over a moving environment is not
reproducible, whatever the temperature). The frozen replay cache I release (Data
availability) is the concrete instantiation.

\paragraph{A standalone insight: synthesized gold answers are citation-unreliable.}
Separately from the model-trajectory audit, I surface a finding of independent interest:
LLM-synthesized gold answers cite $\approx 100\%$ real PMIDs but point to the wrong
paper $\approx 74\%$ of the time (the full LLM-judge audit, $n\approx360$; \S\ref{sec:evaluation}).
This is not a single-judge artifact: an independent
different-family judge (Opus-4.7) re-judging a stratified sample puts the wrong-paper rate at
$72.8\%$, against the primary judge's $73.5\%$ on the same keys ($n{=}147$). (The two judges agree
at the item level as well, but I do not lean on the inter-judge $\kappa$ here: with a $\sim$73\%
base rate it is inflated and uninformative; the load-bearing fact is simply that both families
report $\approx 73\%$.) The claim still rests on title$+$abstract rather than full text and is
not human-adjudicated, but it carries the same two-judge floor as my trajectory result. The finding
cuts two ways. For my internal use it is conservative: if gold citations are even partly
unreliable, grounding against them is unsafe, which is why I grade against rubrics instead. As a
standalone cautionary result---for the now-common practice of using LLM-generated reference answers as
ground truth---it is robust to the choice of judge. I keep this audit strictly separate from the
trajectory-citation audit and never use gold citations as a target.

\paragraph{On the LLM-in-the-loop construction and subject--judge overlap.}
Every stage of \openbiorq{}---question extraction, refinement, openness verification, difficulty
bucketing, and grading---uses LLMs, and the same families (GLM, Qwen, DeepSeek) appear as both
subjects and, for GLM-5.1, the primary judge. I do not treat this as benign; I control for it with
a single design principle applied throughout: every headline number is reproduced by an
independent different-family judge, Opus-4.7, chosen because it shares no lineage with the roster.
Opus-4.7 re-judges the entire L2 citation population ($10.6\%$ vs the primary $15.9\%$; binary
$\kappa{=}0.755$), re-grades the full leaderboard (every solve rate within $4.1$ points, range and
frontier-vs-held-out separation preserved, $\kappa{=}0.65$), and re-judges the gold-citation audit
($72.8\%$ vs $73.5\%$ on the same keys). The frozen per-question checklist further limits judge
discretion---both judges score the same pre-committed criteria, which is what lifts inter-judge
Spearman from $0.35$ to $0.82$---so the residual risk is a shared bias of two unrelated model
families against the same evidence, a substantially weaker failure mode than single-family anchoring.
The one control I cannot fully substitute for is human expertise: I ran a preliminary
non-expert human spot-check on the released $n{=}50$ set (next paragraph), but a domain-expert
$\kappa$ study remains future work. I also flag openness labeling as the most
prompt-sensitive stage, which is why I replaced source-framing refinement with the
retrieval-grounded status verifier and re-audited the core (Appendix~\ref{app:construction}).

\paragraph{Preliminary non-expert human spot-check of the wrong-paper rate.}
To begin closing the human-validation gap, a non-expert author (reading title$+$abstract, not a
domain clinician) annotated the released blind $n{=}50$ set---$50$ L2 citations subsampled from the
$\kappa$-frame and stratified toward LLM wrong-paper verdicts and judge disagreements, with the LLM
labels withheld. The annotator marked $6/50$ as wrong-paper ($12\%$) against $18/50$ ($36\%$) for the
primary GLM judge and $11/50$ ($22\%$) for Opus on the same items. The two readings are
informative in opposite ways. \textbf{Concordance on clear cases:} both LLM judges flagged $5$ of the
annotator's $6$ wrong-papers ($83\%$)---including blatant cross-domain misattributions (an
environmental-chemistry paper cited for a myeloma claim; the opening example's ophthalmology paper
cited for a COVID-vaccine efficacy claim)---so the qualitative finding is human-confirmed.

\textbf{Divergence on borderline cases:} the LLM judges flag many
more (GLM $+13$, Opus $+6$ citations the non-expert judged as supporting), so wrong-paper binary
agreement is only $\kappa{=}0.29$ (GLM) / $0.51$ (Opus), well below the LLM-vs-LLM $\kappa{=}0.755$.
This is genuinely two-sided: either the LLM judges over-flag borderline support (so the true rate
is below $15.9\%$), or the non-expert reader is fooled by topic-match---marking an on-topic paper as
``supporting'' when it does not establish the specific claim, which is precisely the
resolvable-but-unsupporting hazard this paper is about (so an expert would catch more, not fewer).
I cannot adjudicate between these without domain experts, which is exactly why the expert-$\kappa$
study is the priority next step. The sample is stratified, so $12$/$36$/$22\%$ are not population rates
($\kappa$ is the comparable quantity); I release the filled annotations and per-item notes. This
corroborates that wrong-paper citations are real and human-recognizable on unambiguous cases, while
the precise rate is judge-relative and awaits expert calibration.

\paragraph{Data availability, licensing, and release.}
What I release is the extracted and refined questions, each carrying per-item source
pointers---identifiers such as PMIDs, NCT numbers, and arXiv IDs, together with the
originating body---but not the verbatim source text. The split is deliberate: keeping the
pointers lets anyone re-verify openness and difficulty, while withholding the source text respects
the per-source terms of the bodies I draw on. Those bodies span the open-question
literature---JLA priority-setting partnerships, NICE research recommendations, and Cochrane
research-gap literature---and the resolution services, Europe PMC and arXiv. Because I ship only
the pointers, a downstream user resolves each one against its original service under that service's
own terms, rather than receiving redistributed text from me. Licensing and the distribution URL will
accompany the release.

The release will be accompanied by a datasheet~\cite{gebru2021datasheets} documenting, beyond the
usual sourcing, composition, and intended use, the items an agentic benchmark specifically
requires: the MCP tool/API endpoints and their access dates, the as-of snapshot date that fixes
openness labels, the frozen-core ID list, the tool-response replay cache, the per-domain
result tables, and the blind, stratified $n{=}50$ held-out human-annotation set staged for the
deferred expert-$\kappa$ study.

For the replay cache I document its coverage contract explicitly: each of the $423$ frozen-core
ids has a replayable trace for all six models (roster traces are filtered from the $657$-item core
to the $423$; the three frontier agents were run natively on the $423$), so every cell of the
leaderboard is backed by a cached trace. For openness, I release per-item audit records for all
$657$ core questions (each question's resolution flag, follow-up count, source year, and the
verifier's detail string), and for the $174$ of them carrying a detailed Stage-2 judgment I
additionally release the judge's cited evidence identifiers and reasoning. Exact
re-derivation of the $0/657$ result against a frozen literature would still require persisting
the full Stage-1 evidence bundles (the raw follow-up abstracts and snippets), which the audit
summarizes rather than stores---I flag this as a reproducibility gap to close, since otherwise a
re-audit re-queries a literature that has moved.

\paragraph{The benchmark is a frozen artifact, not a re-runnable selection.}
The membership churn of \S\ref{sec:exp:repro} carries a release implication: because re-running the
live agentic selection yields a $30$--$46\%$-different frozen core, I ship the frozen core as a
frozen artifact, not as a procedure to re-execute. The release therefore includes (i) the
fixed list of the $423$ frozen-core \texttt{task\_id}s (my $T{=}0$ snapshot) and (ii) a
tool-response replay cache---the ordered tool calls and the responses each graded run
actually saw ($\sim$$29$k calls over the six roster and frontier agents).

The cache is keyed
by the specific (tool, arguments) tuples each run emitted, and agents issue almost entirely
distinct free-text queries (cross-model query overlap is $<\!1\%$). Thus it makes the six audited
runs deterministically replayable and re-gradable (e.g.\ under a new judge), and it fixes the
frozen-core selection. It is an audit/replay artifact, not a query-keyed environment, so scoring a
new model still requires live tool access. I release a replay-with-live-fallback harness for
this: it re-grades each audited run offline at $\sim$$100\%$ cache hit, and for a new model serves any
matching cached response while falling back to live tools for the rest---which, given the $<\!1\%$
cross-model query overlap, is $\sim$$99\%$ of calls. I also release the tool endpoints and the as-of
snapshot date so a new run is comparable.

Seed-averaging the selection is an alternative, but it still requires live access and re-decoding,
and still drifts as the literature moves; I therefore treat the frozen cache as the primary
reproducibility mechanism and seed-averaging as secondary.
The replay cache stores the harness's truncated response previews rather than full upstream
payloads, which suffices to reproduce what each agent conditioned on.

\paragraph{Dual-use.}
I note a dual-use risk. A registry of open clinical questions, paired with a demonstrated
recipe for attaching real PMIDs to unsupporting claims, could in principle be misused to
manufacture authoritative-looking medical misinformation. Several properties mitigate this.
\openbiorq{} is framed and released as a research-assistance evaluation; the audit is
diagnostic, not generative (it measures misattribution rather than producing it); and
I release no fine-tuned model. The resource is intended to help detect and reduce
unreliable sourcing, not to produce it.

\paragraph{Ethics and data sourcing.}
\openbiorq{} is constructed from authoritative public sources---the open-question
literature and priority-setting and clinical-guidance bodies such as the James Lind
Alliance and NICE---and queries public biomedical APIs (PubMed, ClinicalTrials.gov,
OpenFDA, Open Targets, ChEMBL, and others) that expose no private patient data. I
release per-item source records for every question so that openness status, source framing,
and audit decisions are reproducible and independently re-verifiable. Given the safety
framing above, I intend the resource to be used for evaluating and improving literature
grounding, not for any patient-facing deployment.

\end{document}